\documentclass{article}

\PassOptionsToPackage{numbers, compress}{natbib}
\usepackage[final]{ml4eng_2020}

\usepackage[utf8]{inputenc} %
\usepackage[T1]{fontenc}    %
\usepackage{hyperref}       %
\usepackage{url}            %
\usepackage{booktabs}       %
\usepackage{amsfonts}       %
\usepackage{nicefrac}       %
\usepackage{microtype}      %

\usepackage{amsmath}
\usepackage{graphicx}
\usepackage{subfigure}
\usepackage{algorithmic}
\usepackage{algorithm}
\usepackage{enumitem}
\usepackage{multirow}
\usepackage[title]{appendix}

\title{Combinatorial 3D Shape Generation\\via Sequential Assembly}

\author{%
Jungtaek Kim\\
POSTECH\\
\texttt{jtkim@postech.ac.kr}\\
\And
Hyunsoo Chung\\
POSTECH\\
\texttt{hschung2@postech.ac.kr}\\
\And
Jinhwi Lee\\
POSTECH, POSCO\\
\texttt{jinhwi@postech.ac.kr}\\
\AND
Minsu Cho\\
POSTECH\\
\texttt{mscho@postech.ac.kr}\\
\And
Jaesik Park\\
POSTECH\\
\texttt{jaesik.park@postech.ac.kr}\\
}

\newcommand{\secref}[1]{Section~\ref{#1}}
\newcommand{\figref}[1]{Figure~\ref{#1}}
\newcommand{\tabref}[1]{Table~\ref{#1}}

\newcommand{\algref}[1]{Algorithm~\ref{#1}}

\newcommand{\eqqref}[1]{Eq.~(\ref{#1})}

\newcommand{\argmax}{\operatornamewithlimits{\arg \max}}

\newcommand{\bk}{\mathbf{k}}
\newcommand{\bp}{\mathbf{p}}

\newcommand{\bv}{\mathbf{v}}
\newcommand{\bs}{\mathbf{s}}

\newcommand{\bx}{\mathbf{x}}
\newcommand{\by}{\mathbf{y}}

\newcommand{\bK}{\mathbf{K}}
\newcommand{\bP}{\mathbf{P}}

\newcommand{\calP}{\mathcal{P}}
\newcommand{\calS}{\mathcal{S}}
\newcommand{\calU}{\mathcal{U}}

\newcommand{\bbR}{\mathbb{R}}

\newcommand{\onebyone}{1\!\times\!1}
\newcommand{\twobyfour}{2\!\times\!4}

\newcommand{\eg}{e.g.}
\newcommand{\ie}{i.e.}

\newcommand{\occ}{\textrm{o}}
\newcommand{\sta}{\textrm{s}}

\usepackage{xr}
\makeatletter
\newcommand*{\addFileDependency}[1]{%
  \typeout{(#1)}
  \@addtofilelist{#1}
  \IfFileExists{#1}{}{\typeout{No file #1.}}
}
\makeatother

\begin{document}

\maketitle

\begin{abstract}
Sequential assembly with geometric primitives has drawn attention in robotics and 3D vision 
since it yields a practical blueprint to construct a target shape. 
However, due to its combinatorial property, a greedy method falls short of generating a sequence of volumetric primitives.
To alleviate this consequence induced by a huge number of feasible combinations, 
we propose a combinatorial 3D shape generation framework.
The proposed framework reflects an important aspect of human generation processes 
in real life -- we often create a 3D shape by sequentially assembling unit primitives with geometric constraints.
To find the desired combination regarding combination evaluations, we adopt Bayesian optimization, 
which is able to exploit and explore efficiently the feasible regions constrained by the current primitive placements.
An evaluation function conveys global structure guidance for an assembly process 
and stability in terms of gravity and external forces simultaneously.
Experimental results demonstrate that our method successfully generates combinatorial 3D shapes 
and simulates more realistic generation processes.
We also introduce a new dataset for combinatorial 3D shape generation.
All the codes are available at \url{https://github.com/POSTECH-CVLab/Combinatorial-3D-Shape-Generation}.
\end{abstract}

\section{Introduction\label{sec:intro}}

Constructing a 3D shape via sequential assembly has a huge potential. This generation scheme aims to follow a target shape that can be employed in mimicking a human assembling process and allocating a budget of the number of primitives given. 
In robotics, self-assembly robots~\citep{WeiH2010icra,RomanishinJW2013iros,FeltonSM2013icra,RubensteinM2014science}, inspired by a phenomenon that self-organize a chaotic pattern to an ordered structure in chemistry~\citep{WhitesidesGM2002science} and biology~\citep{HartgerinkJD2001science}, are used in creating a target shape under a specific form of geometric constraints such as contacts and local interactions.
This line of research demonstrates the impressive results that open new applications of sequential assembly. However, these methods have the limitation that they adopt a heuristic or fixed-strategy approach to construct a target shape.

Generic generative models such as variational auto-encoders and generative adversarial networks have been used where the following 3D representation is assumed; a fixed number of points~\citep{AchlioptasP2018icml}, a deformable mesh~\citep{GaoL2019arxiv,GroueixT2018cvpr}, or a voxel grid~\citep{WuJ2016neurips,WuZ2015cvpr}.
However, these generation schemes do not reflect an important aspect of human assembling processes in real life -- we often create a 3D shape by sequentially assembling primitives into a combinatorial configuration.
In this work, we solve a sequential assembling problem with Bayesian optimization-based framework of \emph{combinatorial 3D shape} generation that creates a 3D shape with a set of \emph{geometric primitives}.
Our method can generate a sequence of unit primitives without a significant amount of human effort or a brute-force technique.

In practice, the main challenge lies in a combinatorial explosion as the number of primitives increasing.
For instance, if we assemble $\twobyfour$ LEGO bricks on a free 3D space, the most na\"ive way is to take all combinations into account 
and pick the most probable one for the purpose.
However, with only six $\twobyfour$ LEGO bricks, the number of candidates amounts to 915 million combinations~\citep{EilersS2016amm}.\footnote{Under the assembling conditions suggested in this paper; there exist 46 combinations for two bricks, 3,566 for three bricks, 405,716 for four bricks, and 59,814,648 for five bricks.}
A brute-force approach to combinatorial generation is to find a needle in a haystack due to the prohibitive number of possible combinations.
To tackle this challenge, we propose a sequential assembling method that iteratively evaluates the next possible primitive combinations in a sample-efficient manner, by considering global structure guidance for assembling a target shape and stability in terms of gravity and external forces.
In addition to the proposed pipeline, we introduce a new combinatorial 3D shape dataset that consists of 14 classes and 406 instances.
Due to the nature of the combinatorial shape, the dataset can be readily augmented by manipulating assembling orders.
We hope the new dataset opens a new benchmark for 3D shape generation.

\begin{figure}[t]
    \centering
    \renewcommand{\thesubfigure}{}
    \subfigure[1 step]
    {
        \includegraphics[width=0.14\textwidth]{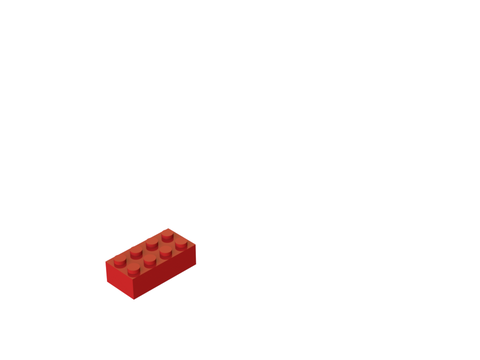}
        \label{fig:teaser_1}
    }
    \subfigure[20 steps]
    {
        \includegraphics[width=0.14\textwidth]{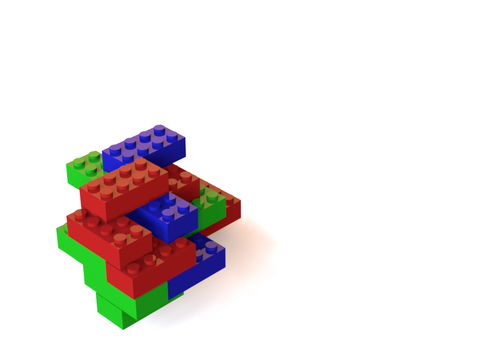}
        \label{fig:teaser_2}
    }
    \subfigure[40 steps]
    {
        \includegraphics[width=0.14\textwidth]{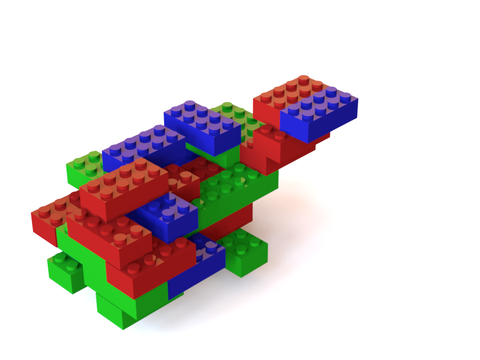}
        \label{fig:teaser_3}
    }
    \subfigure[60 steps]
    {
        \includegraphics[width=0.14\textwidth]{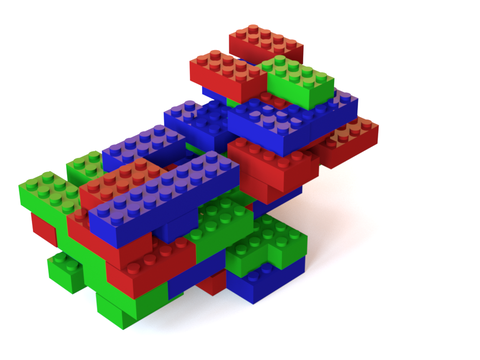}
        \label{fig:teaser_4}
    }
    \subfigure[80 steps]
    {
        \includegraphics[width=0.14\textwidth]{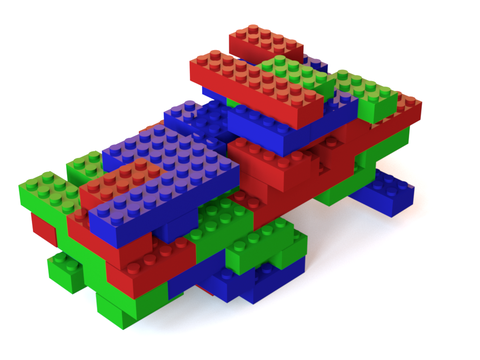}
        \label{fig:teaser_5}
    }
    \subfigure[118 steps]
    {
        \includegraphics[width=0.14\textwidth]{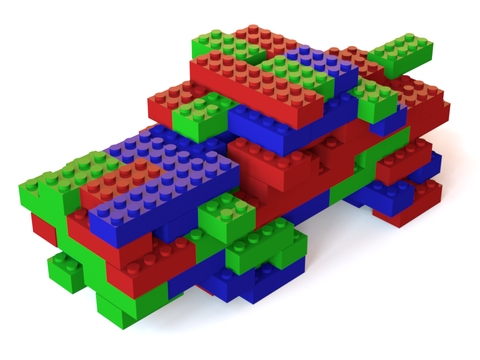}
        \label{fig:teaser_6}
    }
    \caption{Generated assembling sequence that creates a \emph{car} shape with 118 unit primitives.}
    \label{fig:teaser}
\end{figure}

\section{Related Work\label{sec:related}}

\paragraph{Self-Assembly Robots}
This class of robots has been inspired by a self-assembly phenomenon of the nature~\citep{WhitesidesGM2002science,HartgerinkJD2001science}, 
which is driven by physical interaction between molecules or unit components, 
or surrounding environments.
\citet{WeiH2010icra} present a self-configurable and self-assembly modular robot, 
the actuator of which can enable itself to move and dock to other modules.
\citet{RomanishinJW2013iros} propose a magnetic modular robot, which can 
move independently with the flywheels installed in the edges of robot.
\citep{FeltonSM2013icra} suggests a printed self-folding robot that can be used in various 
fields such as micro-assembly and space applications.
\citet{RubensteinM2014science} propose a flock of programmable self-assembly robots, 
which can create a target shape without external intervention.

\paragraph{Topology and Layout Optimization}
Topology optimization~\citep{BendsoeMP1988CMAME}, which finds an optimal layout 
where predefined configurations and constraints are provided, 
has widely been used in shape design, prototyping, and manufacturing.
\citet{EschenauerHA1994so} introduce a method that inserts holes into a component with iterative positioning.
\citet{BorrvallT2003ijnmf} use a topology optimization technique in fluid mechanics to solve applications such as pipe and airfoil designs.
\citet{KharmandaG2004smo} suggest a method to find reliable and efficient structures with the reliability index.
\citet{BrackettD2011sffs} utilize topology optimization in additive manufacturing for producing end-use parts.
Moreover, to find an optimal layout for 3D objects, 
\citet{TestuzR2013eurographics} identify a suitable primitive set 
for a given mesh and applies a greedy method to repair weak connections.
\citet{LeeS2015gecco} propose to optimize a primitive layout using genetic algorithm.
\citet{LuoSJ2015acmtgraphics} consider the physical stability of constructed model, which helps to create realistic and realizable assembly accomplishments.

\paragraph{Generative Models for 3D Objects}
\citet{AchlioptasP2018icml} learn representations from point clouds via autoencoder.
Their approach employs either raw point sets or learned representations, 
to train a generative adversarial network.
For the deformable mesh generation, \citet{GroueixT2018cvpr} suggest a method 
to transform 2D texture map atlases into 3D surface.
\citet{GaoL2019arxiv} generate structured deformable meshes.
The network is composed of part-level and structural-level variational autoencoder.
On the other hand, convolutional deep belief networks~\citep{WuZ2015cvpr} 
and generative adversarial networks~\citep{WuJ2016neurips} are used to generate an occupancy grid.
\citet{NashC2020icml} predict mesh vertices and faces using Transformer-based neural networks.

Compared to the aforementioned techniques that attempt to find an optimal layout 
in terms of their own evaluation metrics, our goal is to create a sequence of 
unit primitives, considering the stability of combination as well as following a desired shape.
To the best of our knowledge, this is the first attempt to build a 3D shape 
via a sequential and combinatorial approach.
In particular, our algorithm efficiently seeks a feasible primitive combination 
using Bayesian optimization, by reducing the number of observations required.

\section{Sequential Assembly with Unit Primitives\label{sec:formulation}}

In this section, we provide the detailed configurations and assumptions, 
which are used to propose our combinatorial 3D shape generation method.
Volumetric representation of our interest is the most straightforward expression 
to define interactions between primitives 
and assemble a shape with a set of geometric primitives.
However, the choice for the types of unit geometric primitives is important, 
because it makes this sequential assembly either too simple or too challenging.

\begin{figure}[t]
    \centering
    \subfigure[Target shape]
    {
        \includegraphics[width=0.21\textwidth]{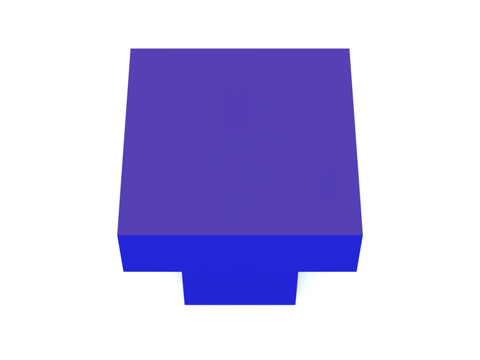}
        \label{fig:motivation_0}
    }
    \subfigure[$\onebyone$-sized primitives]
    {
        \includegraphics[width=0.21\textwidth]{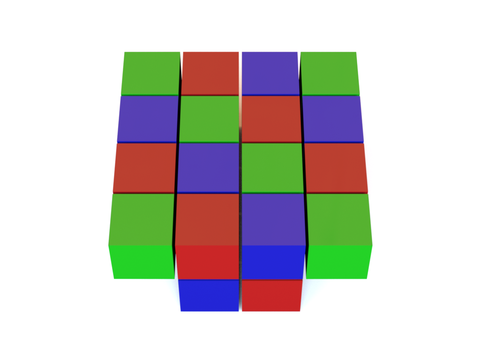}
        \label{fig:motivation_1}
    }
    \subfigure[$\twobyfour$-sized primitives]
    {
        \includegraphics[width=0.21\textwidth]{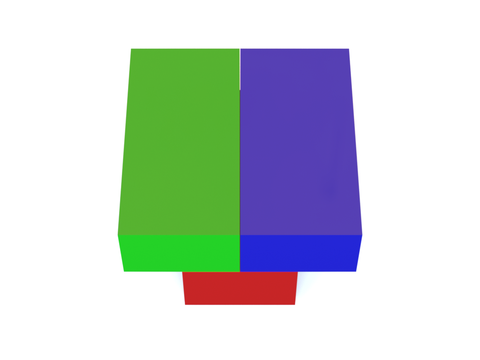}
        \includegraphics[width=0.21\textwidth]{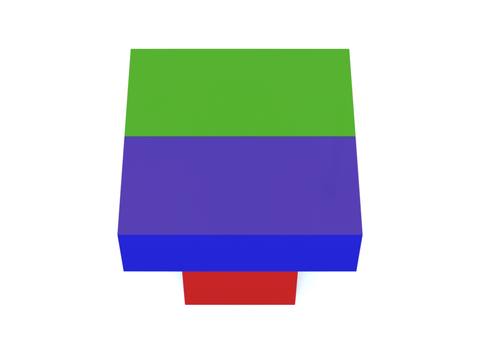}
        \label{fig:motivation_2}
    }
    \caption{Results on assembling a target shape with $\onebyone$-sized primitives and $\twobyfour$-sized primitives.}
    \label{fig:motivation}
\end{figure}

Assume that we have a target shape (\ie, \figref{fig:motivation_0}) to assemble.
If we use $\onebyone$-sized voxel primitives, there exists only \emph{one combination} 
composed of 24 primitives, as shown in \figref{fig:motivation_1}.
However, if we use the primitives identical to $\twobyfour$ volumetric primitives, 
we can assemble \emph{two combinations} (\ie, \figref{fig:motivation_2}).
These observations indicate that smaller and simpler primitives tend to create more fine-grained shapes 
but have less combinatorial sequences.
On the contrary, larger and more complicated primitives tend to create coarser shapes 
but have more interesting combinatorial sequences.
These facts also relates to the real-world problem when we realize shapes by assembling multiple parts.
As a result, we restrict the primitive we employ as a single type, 
which can only be connected with another primitive 
in fixed configurations 
and does not allow overlap between primitives.
In this work, we select a $\twobyfour$ LEGO brick as a unit primitive, since a $\twobyfour$-sized primitive with studs and cavities is one of the representative basic LEGO bricks.\footnote{We also have a historical reason for this choice. As introduced in \citep{HermanS2012book}, a $\twobyfour$ LEGO brick is the first patent granted in 1947 of the LEGO company and it opens a progressive development of building a 3D shape.}

\section{Occupiability Grid\label{sec:grid}}

To tackle an assembly problem, we start by defining a search region $\calS \subseteq \bbR^3$, 
which is a space to construct a 3D object. 
We employ \emph{occupiability grid}, which is the opposite concept of probabilistic volumetric models~\citep{PollardT2007cvpr}.
The occupiability grid is a grid of which the unit cell (\ie, a voxel) possesses the possibility of being occupied in the future.

Given the number of partitions for each of three axes $m_1$, $m_2$, and $m_3$, 
a voxel $\bv_{ijk}$ can be represented as $(i, j, k)$
where $i \in [m_1]$, $j \in [m_2]$, and $k \in [m_3]$,
and a collection of entire voxels is $\{ \bv_{ijk} \}_{i \in [m_1], j \in [m_2], k \in [m_3]}$.
For a generic voxel grid, the occupancy of a voxel is expressed as one of two alternatives:
\begin{equation}
    q(\bv_{ijk}) = \left\{ \begin{array}{cr}
        1 & \quad \textrm{if it is occupied,} \\
        0 & \textrm{otherwise.}
    \end{array} \right.
    \label{eqn:voxel}
\end{equation}

On the other hand, the occupiability of voxel $\bv_{ijk}$ can be expressed as
\begin{equation}
    o(\bv_{ijk}) = \left\{ \begin{array}{cr}
        1 & \quad \textrm{if it is occupiable,} \\
        0 & \textrm{otherwise.}
    \end{array} \right.
    \label{eqn:occupiability}
\end{equation}
In contrast with voxel definition, our concept of occupiability is to capture the likelihood of an empty voxel would be occupied in the near future. 
Thus, the voxel that is already occupied or prohibited due to specific constraints 
will be regarded as 0 occupiability.

Since a volumetric $\twobyfour$ brick as a unit primitive is placed in $\calS$, it should transform into a single covariate to compare to other primitives.
We thus denote the coordinate of each primitive as a 3D vector $\bx = [x_1, x_2, x_3] \in \calS$, where $(x_1, x_2)$ is the center over the first two axes and $x_3$ is the bottom of the primitive, and the direction of each primitive as a scalar $d$.
With this representation, suppose that every primitive is placed over the plane that $x_3 = 0$, and $d$ of each primitive is placed either lengthwise (\ie, denoted as $d = 0$) or breadthwise (\ie, denoted as $d = 1$).
It means every primitive can turn either $0$, $\pi/2$, $\pi$, or $3\pi/2$ radians along the third axis of $\calS$.
To sum up, each primitive is defined as a tuple $(\bx, d)$ (henceforth, denoted as $\bp \in \calP$ where $\calP = \calS \times \{0, 1\}$)
where $x_3 \geq 0$ and $d \in \{0, 1\}$, and a $n$-primitive combination is expressed as a set $\{ \bp_i \}_{i = 1}^n = \{ (\bx_i, d_i) \}_{i = 1}^n$.

\section{Combinatorial 3D Shape Generation\label{sec:method}}

\begin{figure}[t]
    \centering
    \includegraphics[width=0.98\textwidth]{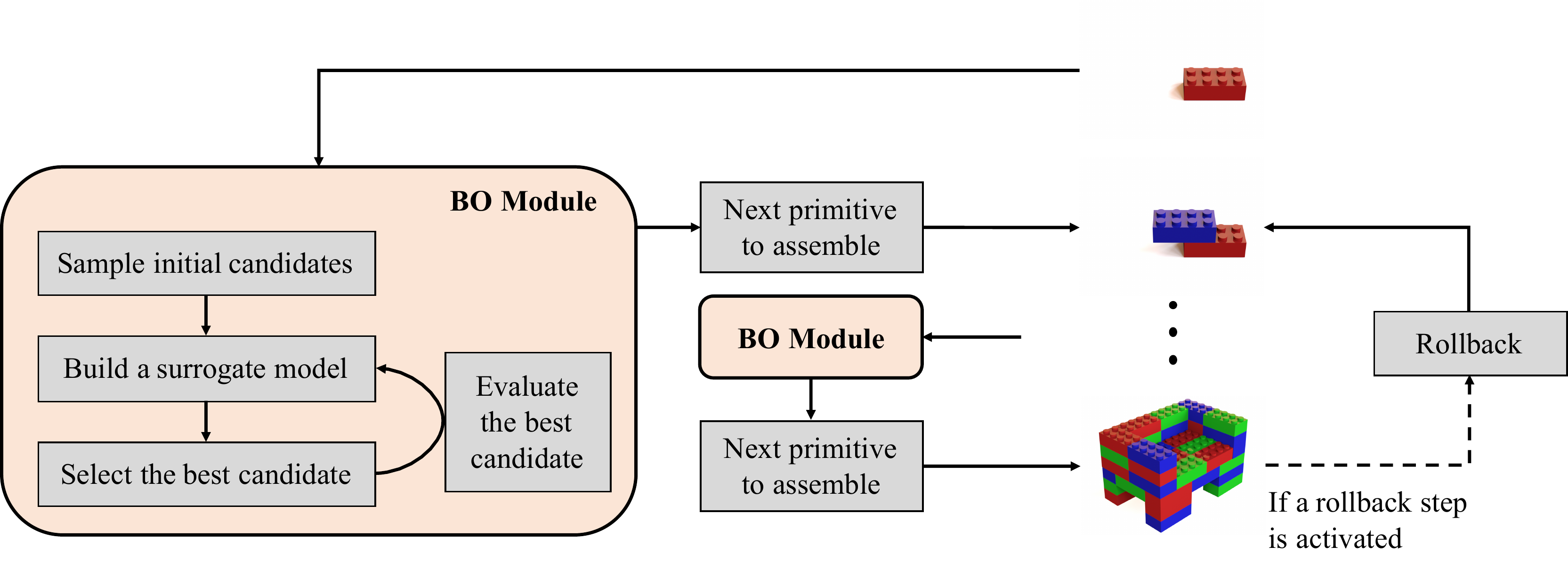}
    \caption{Overview of our combinatorial 3D shape generation pipeline.}
    \label{fig:overview}
\end{figure}

An assembling sequence is generated where each step in the series 
is suggested by one of the sequential model-based optimization methods, 
Bayesian optimization~\citep{BrochuE2010arxiv}.
The Bayesian optimization strategy efficiently samples the position 
of the next primitive to assemble.

\paragraph{Evaluating Primitive Combinations}
To determine the position of the next primitive guided by Bayesian optimization,
we need to define two evaluation functions over $(n + 1)$-primitive combinations.
One $f_{\occ}$ is related to occupiability and the other $f_{\sta}$ is related to stability:
\begin{equation}
    y_{\occ} = f_{\occ} \left(\bp_{n + 1}; \{ \bp_l \}_{l = 1}^n \right) + \epsilon_{\occ}
    \quad \textrm{and} \quad
    y_{\sta} = f_{\sta} \left(\bp_{n + 1}; \{ \bp_l \}_{l = 1}^n \right) + \epsilon_{\sta}, \label{eqn:scores_occ_sta}
\end{equation}
which are considered as true functions, where $\epsilon_{\occ}$ and $\epsilon_{\sta}$ are observation noises.

We design $f_{\occ}$ to guide to follow a global shape context 
without providing exact probable positions of the primitive we would like to place.
$f_{\sta}$ is to measure time to be stable 
where gravity is existed and external forces at the beginning are applied.
Suppose that $\bv_{ijk}$ is a voxel of our interest and $\{ \bp_l \}_{l = 1}^n$ is the current primitive combination.
Given a desired shape $\{\bv_l\}_{l = 1}^{N}$, 
the occupiability is the possibility to be occupied:
\begin{equation}
    o(\bv_{ijk}; \{ \bp_l \}_{l = 1}^n)
    = q\left(\bv_{ijk}; \{\bv_l\}_{l = 1}^N \right)
    \wedge \left( 1 - q\left(\bv_{ijk}; \{\bp_l\}_{l = 1}^n \right) \right).
    \label{eqn:prob_occupiability}
\end{equation}
With \eqqref{eqn:scores_occ_sta} and \eqqref{eqn:prob_occupiability}, the occupiability score over $\bp_{n + 1}$ is
\begin{equation}
	f_{\occ} \left(\bp_{n + 1}; \{ \bp_l \}_{l = 1}^n \right) = \sum_{\bv_{ijk} \in \bp_{n + 1}} o(\bv_{ijk}; \{ \bp_l \}_{l = 1}^n).
\end{equation}

As the other evaluation function, we measure time to be stable 
where gravity is constantly applied to the combination and 
external forces are applied at the beginning of the simulation.
Since measuring time to be stable in practice is a time-consuming and difficult task, 
we test an artificial experiment on a physics engine simulator.
To precisely measure the stability where instability is implanted,
we apply the external forces to one of four directions alternately at the beginning, 
and then measure time steps until the position of combinations is not changed:
$f_{\sta} \left(\bp_{n + 1}; \{ \bp_l \}_{l = 1}^n \right) = \tau$, 
if $\frac{1}{n + 1} \sum_{l = 1}^{n + 1} \| \bp_l^{(\tau)} - \bp_l^{(\tau - 1)} \|_2 < \varepsilon$,
where $\bp^{(\tau)}$ indicates the primitive position at a time step $\tau$, 
and $\varepsilon$ is a threshold for terminating the simulation.
The details are described in \secref{sec:exp} and the appendices.

\begin{algorithm}[t]
	\caption{Find a Feasible Primitive Combination}
	\label{alg:combination}
	\begin{algorithmic}[1]
		\REQUIRE An initial primitive combination $\{\bp_i\}_{i = 1}^m$,
			the number of primitives to assemble $T$,
			the number of rollback steps $\tau$,
			threshold for rolling back $\alpha$.
		\ENSURE A primitive combination $\{\bp_{j}\}_{j = 1}^{m + T}$.
		\STATE Initialize $t \leftarrow 0$ and compute $y_{\occ, 0}$, $y_{\sta, 0}$ with $\{\bp_t\}_{i = 1}^{m}$.
		\WHILE {$t < T$}
			\STATE Select a next primitive to assemble $\bp_{t + 1}$, using \algref{alg:select}.
			\STATE Assemble $\bp_{t + 1}$ to the combination, $\{\bp_t\}_{i = 1}^{m + t}$.
			\STATE Update $t \leftarrow t + 1$.
			\STATE Compute $y_{\occ, t}$, $y_{\sta, t}$ with $\{\bp_t\}_{i = 1}^{m + t}$.
			\IF {$t \geq \tau$ and $\sum_{k = 0}^{\tau - 1} \left( \max_{\bp_{m + t - k}} y_{\occ, t - k} \right) - y_{\occ, t - k} < \alpha$}
			    \STATE Roll back $\{\bp_t\}_{i = 1}^{m + t}$ to $\{\bp_t\}_{i = 1}^{m + t'}$ where $t' = t - \tau$.
			    \STATE Update $t \leftarrow t - \tau$.
            \ENDIF
		\ENDWHILE
		\STATE \textbf{Return} $\{\bp_{j}\}_{j = 1}^{m + T}$
	\end{algorithmic}
\end{algorithm}

\begin{algorithm}[t]
	\caption{Query a Candidate of Next Primitive}
	\label{alg:query}
	\begin{algorithmic}[1]
		\REQUIRE A $n$-primitive combination $\{ \bp_i \}_{i = 1}^n$,
			$r$ possible primitives observed and their evaluation scores with the $n$-primitive combination $\{(\bp_j, s_j)\}_{j = 1}^r$,
			the number of samples for acquisition function optimization $\zeta$.
		\ENSURE A candidate of next primitive $\bp_{r + 1}^*$.
		\STATE Estimate a surrogate function:
		\begin{equation}
			\hat{f} \left( \bp; \{\bp_i\}_{i = 1}^n, \{(\bp_j, s_j)\}_{j = 1}^r \right),\nonumber
		\end{equation}
		using the primitive combination and the historical observations.
		\STATE Sample $\zeta$ primitives possible to assemble.
		\STATE Find a maximizer $\bp_{r + 1}^*$, \ie, one of $\zeta$ primitives, of the acquisition function computed by $\hat{f}$.
		\STATE \textbf{Return} $\bp_{r + 1}^*$
	\end{algorithmic}
\end{algorithm}

\begin{algorithm}[t]
	\caption{Select the Next Primitive Position to Assemble}
	\label{alg:select}
	\begin{algorithmic}[1]
		\REQUIRE Initial primitive combination $\{\bp_i\}_{i = 1}^m$,
			the number of initial primitives $v$,
			the number of primitive candidates $q > v$.
		\ENSURE The next primitive to assemble $\bp_{m + 1}$.
		\STATE Sample $v$ primitives to be able to assemble randomly, $\{\bp_i\}_{i = 1}^v$.
		\STATE Evaluate the primitive combinations each of which is composed of initial combination $\{\bp_i\}_{i = 1}^m$ and one of $v$ primitives, $\{ (\bp_i, y_{\occ, i}, y_{\sta, i}) \}_{i = 1}^v$.
		\FOR {$j = v+1, \ldots, q$}
			\STATE Query the next primitive candidate to assemble $\bp_{j}$.
			\STATE Evaluate the primitive combination composed of $\{\bp_i\}_{i = 1}^m$ and $\bp_{j}$.
			\STATE Update the candidate set, $\{(\bp_i, y_{\occ, i}, y_{\sta, i})\}_{i = 1}^j$.
		\ENDFOR
			\STATE Select the next primitive $\bp_{m + 1}$, which has achieved the best score from $\{ (\bp_i, y_{\occ, i}, y_{\sta, i}) \}_{i = 1}^{q}$.
		\STATE \textbf{Return} $\bp_{m + 1}$
	\end{algorithmic}
\end{algorithm}

\paragraph{Determining the Next Primitive to Assemble}
Using the aforementioned evaluation functions over primitive combinations, 
we can efficiently determine the next position to assemble.
Bayesian optimization, which is a sample-efficient global optimization method for black-box functions, 
sequentially finds the next primitive candidate that maximizes an acquisition function.
The benefit of utilizing Bayesian optimization is 
that we do not need to assume differentiability, continuity, 
or any other specific functional form 
of the original function~\citep{BrochuE2010arxiv,ShahriariB2016procieee}.

As shown in \eqqref{eqn:scores_occ_sta}, 
the evaluation functions that define where we should assemble cannot be optimized 
using generic optimization strategies due to the unknown of functional forms.
For this property, Bayesian optimization is utilized to decide 
where a primitive should be assembled without human intervention.
Moreover, determining a primitive position to assemble is taken into account 
as a process to reveal where we assemble the next position among huge possible primitive positions,
which is a sequential combinatorial procedure to assembling primitives with Bayesian optimization.
In this intuition, \algref{alg:combination}, \algref{alg:query}, 
and \algref{alg:select} are introduced.

First of all, similar to common Bayesian optimization~\citep{JonesDR98jgo,MockusJ1978tgo}, 
a surrogate function over primitives is built given $r$ historical observations $\{ (\bp_i, y_{\occ, i}, y_{\sta, i}) \}_{i = 1}^r$.
Commonly, Gaussian process regression~\citep{RasmussenCE2006book} is used as a surrogate function 
in the Bayesian optimization community~\citep{BrochuE2010arxiv,SnoekJ2012neurips},
because it can express any function in the reproducing kernel Hilbert space.
Note that each primitive $\bp$ is regarded as a four-dimensional vector, 
composed of a 3D vector of primitive $\bx$ and a direction of primitive $d$.
By the Gaussian process regression,
given $r$ four-dimensional inputs
$\bP = [\bp_1 \cdots \bp_r]^\top \in \bbR^{r \times 4}$, 
their associated outputs for occupiability $\by_{\occ} = [y_{\occ, 1} \cdots y_{\occ, r}] \in \bbR^{r}$, 
and their associated outputs for stability $\by_{\sta} = [y_{\sta, 1} \cdots y_{\sta, r}] \in \bbR^{r}$, 
a function value and its uncertainty are represented 
by the posterior mean and variance functions for occupiability or stability: 
$\mu_j(\bp) = \bk(\bp, \bP) \left( \bK(\bP, \bP) + \sigma_n^2 \boldsymbol I \right)^{-1} \by_j$ 
and $\sigma^2_j(\bp) = k(\bp, \bp) - \bk(\bp, \bP) \left( \bK(\bP, \bP) + \sigma_n^2 \boldsymbol I \right)^{-1} \bk(\bP, \bp)$, 
where $j \in \{ \occ, \sta \}$, and 
the covariance functions are defined as 
$k: \bbR^4 \times \bbR^4 \to \bbR$, 
$\bk: \bbR^4 \times \bbR^{r \times 4} \to \bbR^r$, 
and $\bK: \bbR^{r_1 \times 4} \times \bbR^{r_2 \times 4} \to \bbR^{r_1 \times r_2}$.
In addition, $\sigma_n^2$ is a noise scale and 
$\boldsymbol I$ is an identity matrix.

We compute the acquisition function values for $\zeta$ primitives possible to assemble, 
and find a maximizer among the $\zeta$ primitives.
To cope with two evaluation functions, 
we employ a multi-objective Bayesian optimization with random scalarizations~\citep{PariaB2019uai}.
A maximizer $\bp^*$ of the acquisition function over $\bp$ 
is found to observe the true functions:
\begin{equation}
    \bp^* = \argmax_{\bp \in \calP} \lambda_{\occ} a(\bp; \mu_{\occ}(\bp), \sigma_{\occ}^2(\bp)) + \lambda_{\sta} a(\bp; \mu_{\sta}(\bp), \sigma_{\sta}^2(\bp)),
    \label{eqn:max_acq}
\end{equation}
where $\calP$ is a compact set, and 
$\lambda_{\occ}$, $\lambda_{\sta}$ are random weights sampled from uniform distributions.
As an acquisition function, in this paper, 
we use Gaussian process upper confidence bound (GP-UCB)~\citep{SrinivasN2010icml}, 
$a_{\textrm{UCB}}(\bp; \bP, \bs) = \mu(\bp) + \gamma \sigma(\bp)$,
where $\gamma$ is a trade-off hyperparameter for exploitation and exploration.
This procedure is presented in \algref{alg:query}.

Sampling $\zeta$ is the main difference 
between our method and common Bayesian optimization strategies.
Well-known techniques for optimizing an acquisition function 
(\eg, DIRECT~\citep{JonesDR1993jota} and L-BFGS-B~\citep{LiuDC1989mp}) poorly work,
because our search space contains the complicated constraints 
that are basically determined by occupiabilities.
Besides, the combinatorial approach~\citep{BaptistaR2018icml} is difficult 
to apply due to the curse of dimensionality, which is derived 
from the combinatorial explosion of inputs.
We thus sample a feasible set from a primitive set, each element of which can assemble.
This technique is used in the automated machine learning community~\citep{HutterF2011lion}.

As shown in \algref{alg:select}, after choosing $v$ initial random primitives 
and evaluating those primitives with $\{ \bp_i \}_{i = 1}^m$, 
where $m$ is the cardinality of given primitive combination, a primitive candidate is queried, 
and new observation is accumulated, until $q$ primitives are observed.
Finally, the next primitive that has achieved the best score for occupiability is returned.
\algref{alg:combination} describes how a feasible primitive combination is assembled 
with the evaluation functions that guide an assembling process.
Consequently, we obtain a $(m + T)$-primitive combination.

\begin{figure}[t]
    \centering
    \renewcommand{\thesubfigure}{}
    \subfigure[(Cup) 10 steps]
    {
        \includegraphics[width=0.18\textwidth]{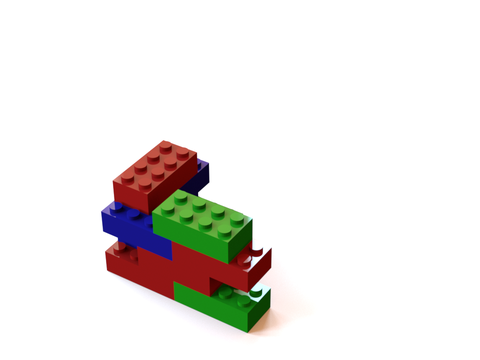}
        \label{fig:cup_1}
    }
    \quad
    \subfigure[20 steps]
    {
        \includegraphics[width=0.18\textwidth]{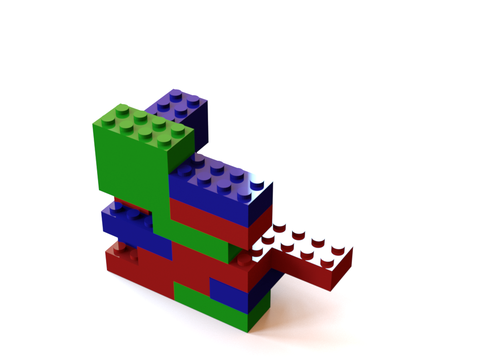}
        \label{fig:cup_2}
    }
    \quad
    \subfigure[30 steps]
    {
        \includegraphics[width=0.18\textwidth]{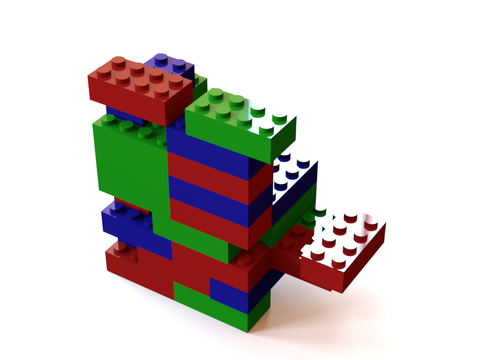}
        \label{fig:cup_3}
    }
    \quad
    \subfigure[55 steps]
    {
        \includegraphics[width=0.18\textwidth]{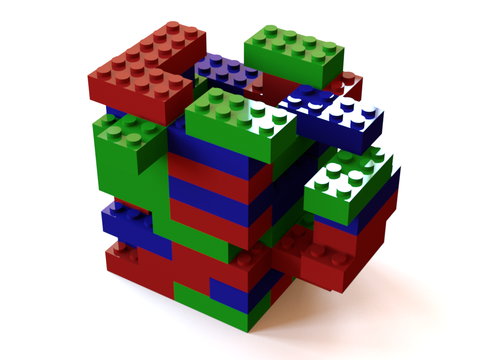}
        \label{fig:cup_4}
    }
    \\
    \renewcommand{\thesubfigure}{}
    \subfigure[(Bench) 15 steps]
    {
        \includegraphics[width=0.18\textwidth]{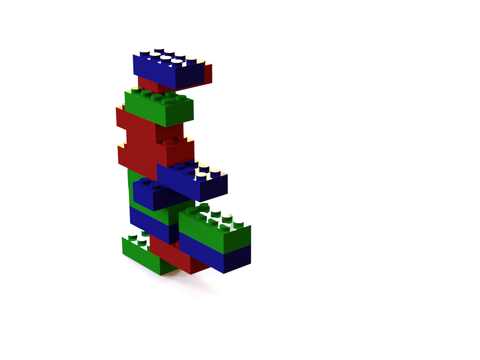}
        \label{fig:bench_1}
    }
    \quad
    \subfigure[30 steps]
    {
        \includegraphics[width=0.18\textwidth]{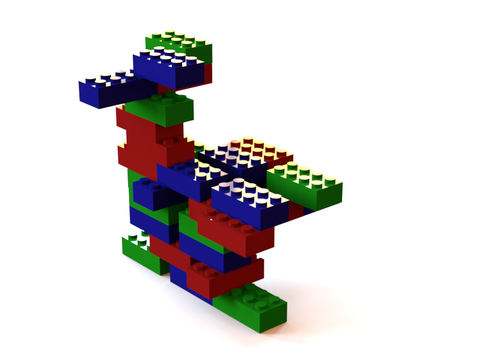}
        \label{fig:bench_2}
    }
    \quad
    \subfigure[45 steps]
    {
        \includegraphics[width=0.18\textwidth]{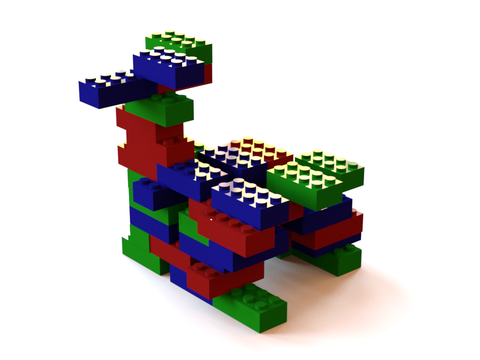}
        \label{fig:bench_3}
    }
    \quad
    \subfigure[70 steps]
    {
        \includegraphics[width=0.18\textwidth]{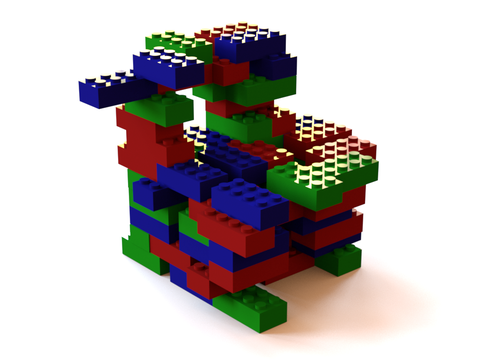}
        \label{fig:bench_4}
    }
    \\
    \renewcommand{\thesubfigure}{}
    \subfigure[(Table) 5 steps]
    {
        \includegraphics[width=0.18\textwidth]{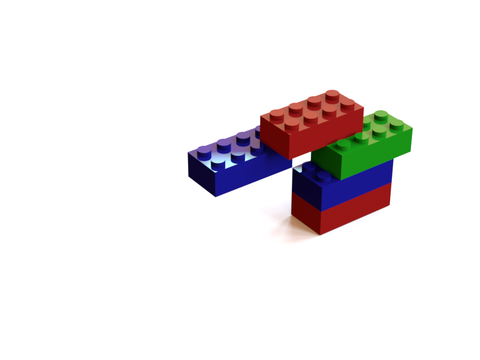}
        \label{fig:table_1}
    }
    \quad
    \subfigure[10 steps]
    {
        \includegraphics[width=0.18\textwidth]{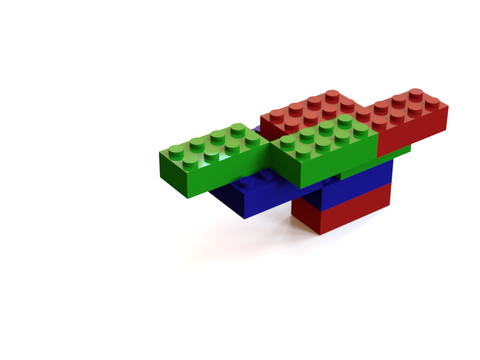}
        \label{fig:table_2}
    }
    \quad
    \subfigure[15 steps]
    {
        \includegraphics[width=0.18\textwidth]{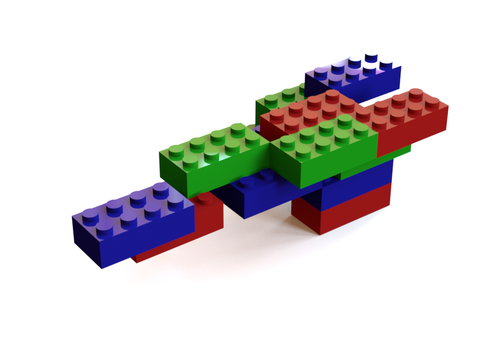}
        \label{fig:table_3}
    }
    \quad
    \subfigure[29 steps]
    {
        \includegraphics[width=0.18\textwidth]{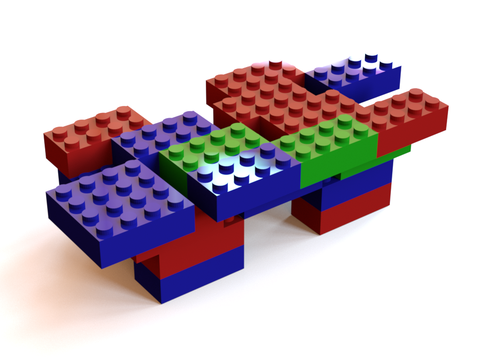}
        \label{fig:table_4}
    }
    \\
    \renewcommand{\thesubfigure}{}
    \subfigure[(Sofa) 15 steps]
    {
        \includegraphics[width=0.18\textwidth]{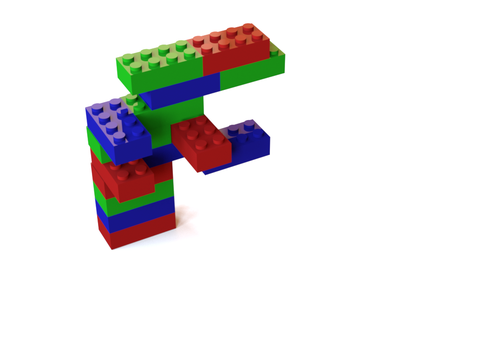}
        \label{fig:sofa_1}
    }
    \quad
    \subfigure[30 steps]
    {
        \includegraphics[width=0.18\textwidth]{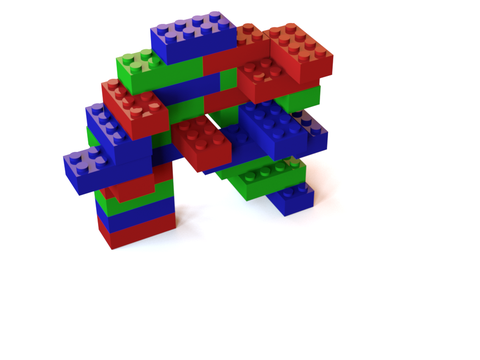}
        \label{fig:sofa_2}
    }
    \quad
    \subfigure[45 steps]
    {
        \includegraphics[width=0.18\textwidth]{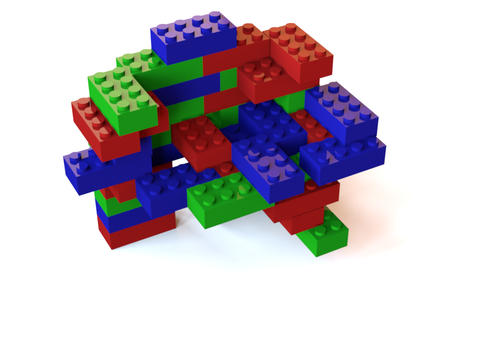}
        \label{fig:sofa_3}
    }
    \quad
    \subfigure[64 steps]
    {
        \includegraphics[width=0.18\textwidth]{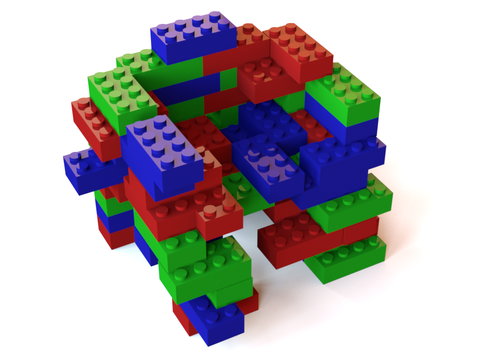}
        \label{fig:sofa_4}
    }
    \caption{Results on sequential primitive assemblies.}
    \label{fig:assembly_cup_bench}
\end{figure}

\begin{figure}[t]
    \centering
    \subfigure[Height]
    {
        \includegraphics[width=0.20\textwidth]{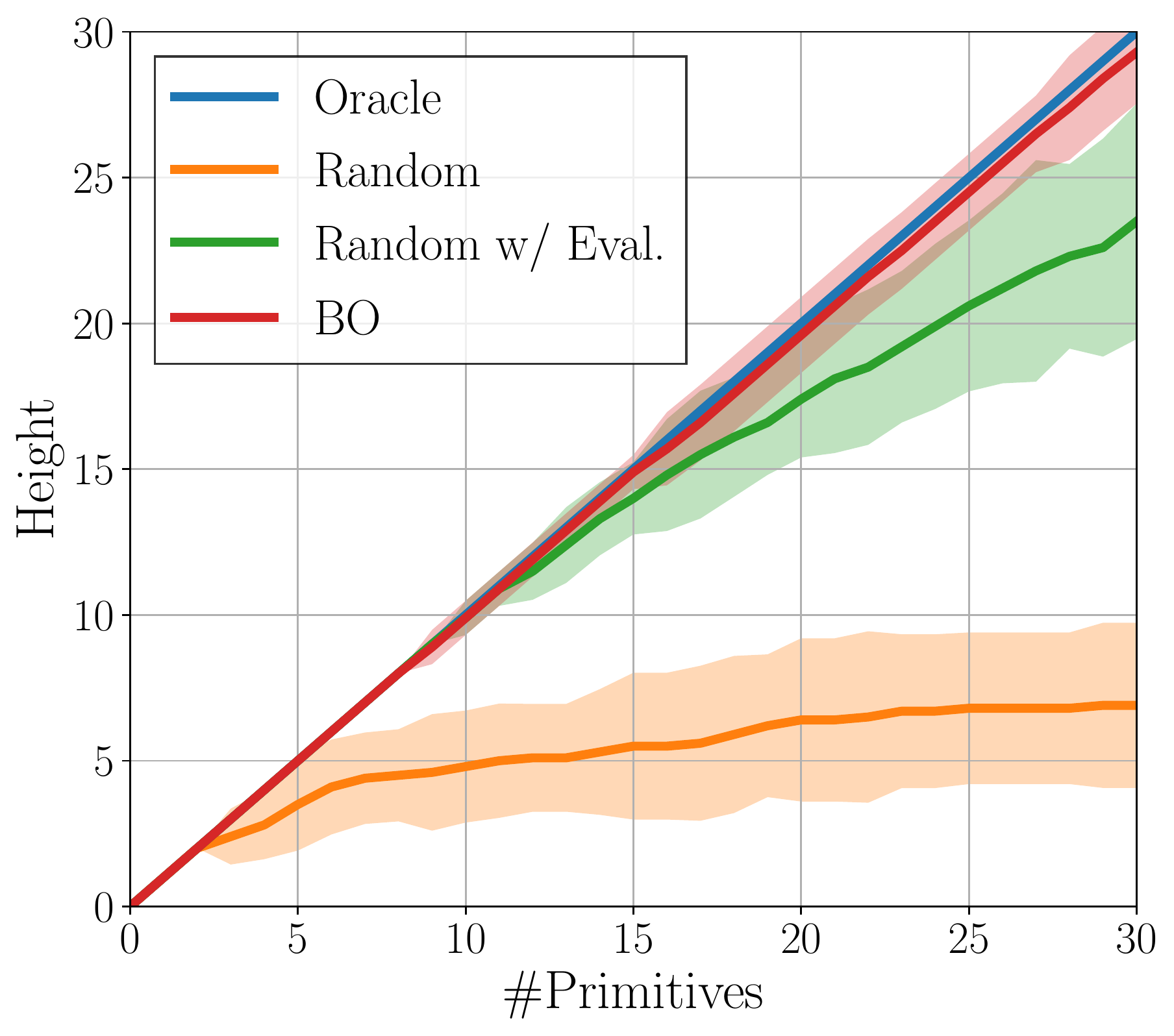}
        \label{fig:height}
    }
    \subfigure[Width]
    {
        \includegraphics[width=0.20\textwidth]{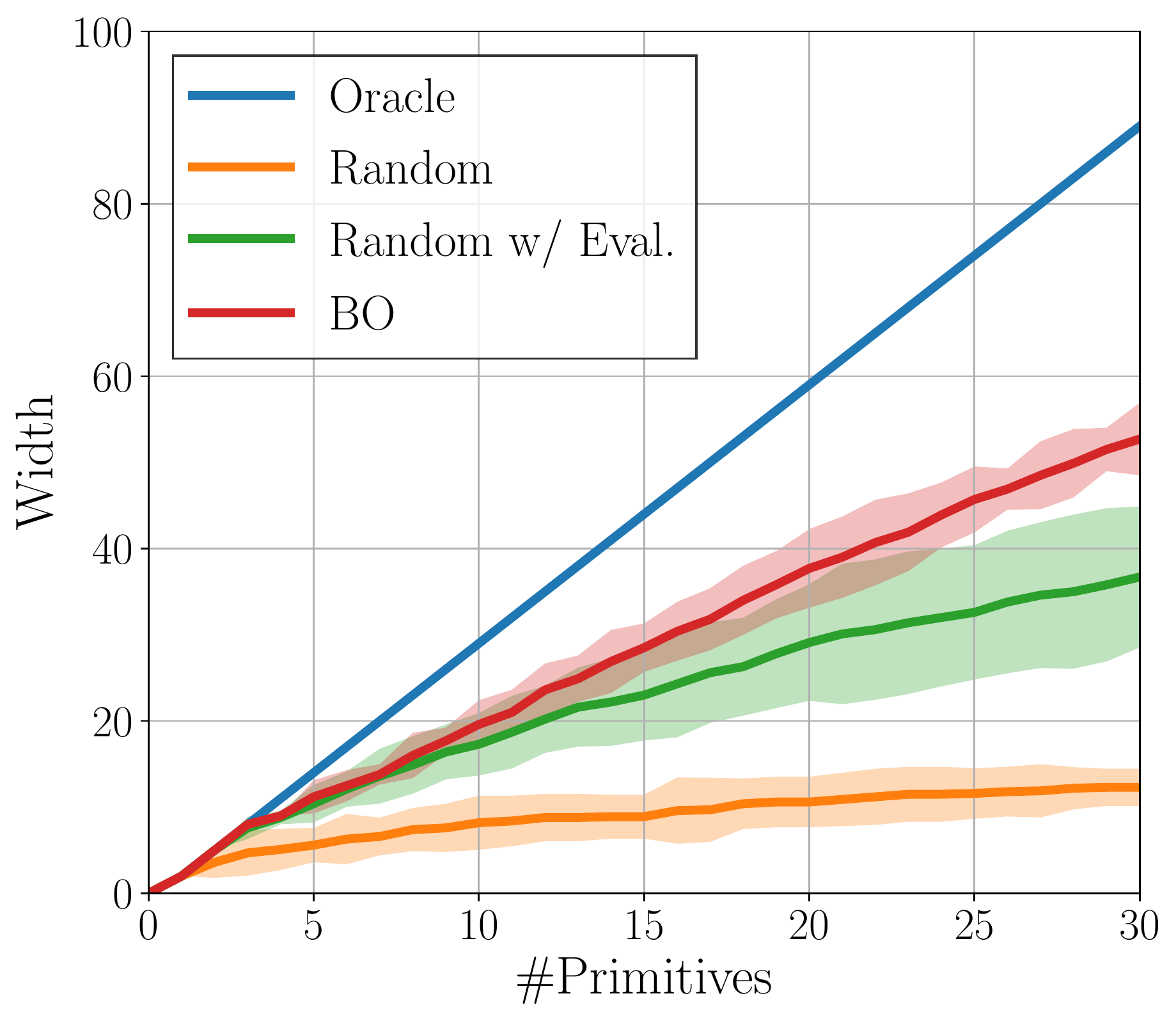}
        \label{fig:width}
    }
    \subfigure[Depth]
    {
        \includegraphics[width=0.20\textwidth]{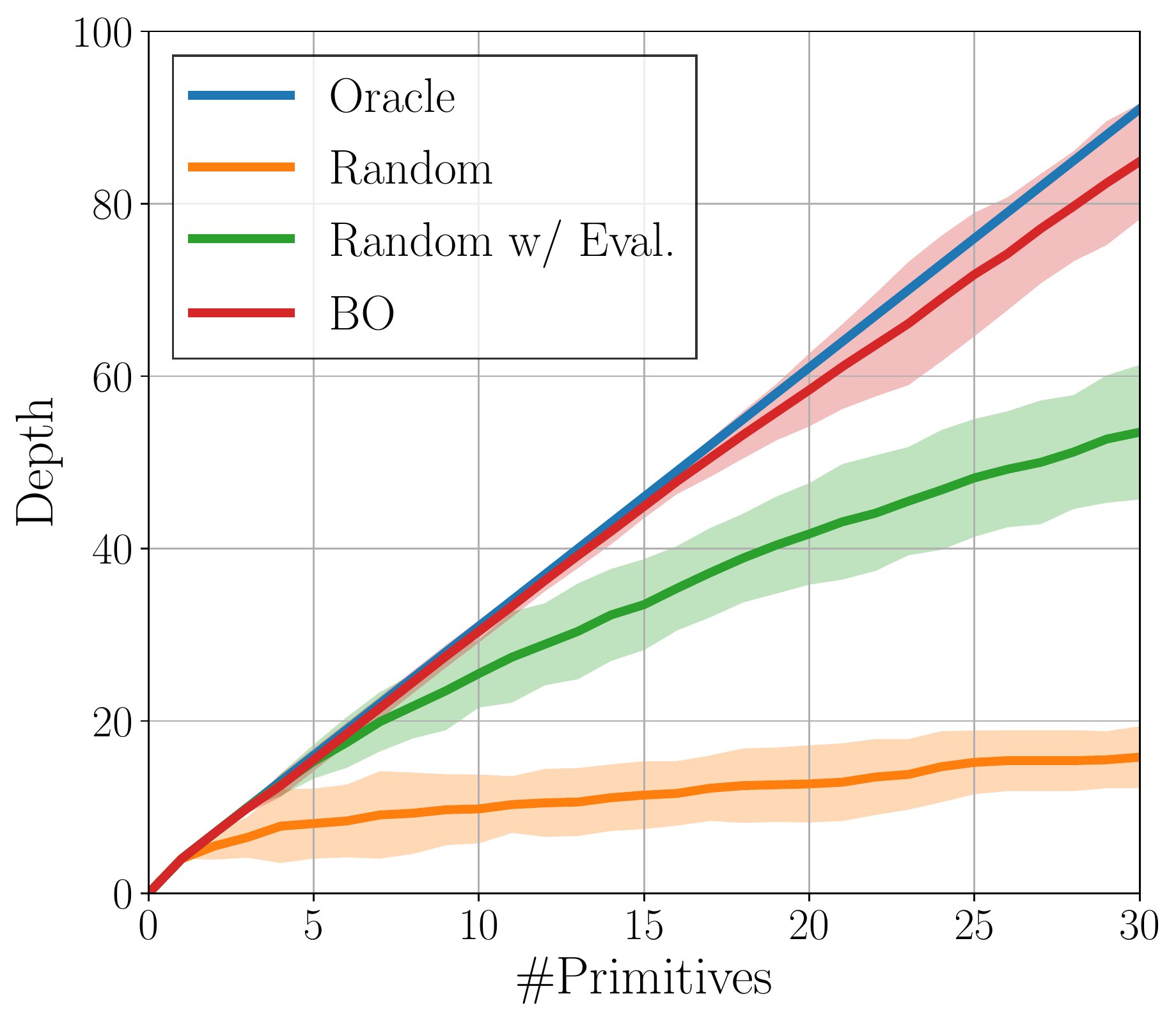}
        \label{fig:depth}
    }
    \subfigure[\#Connected studs]
    {
        \includegraphics[width=0.20\textwidth]{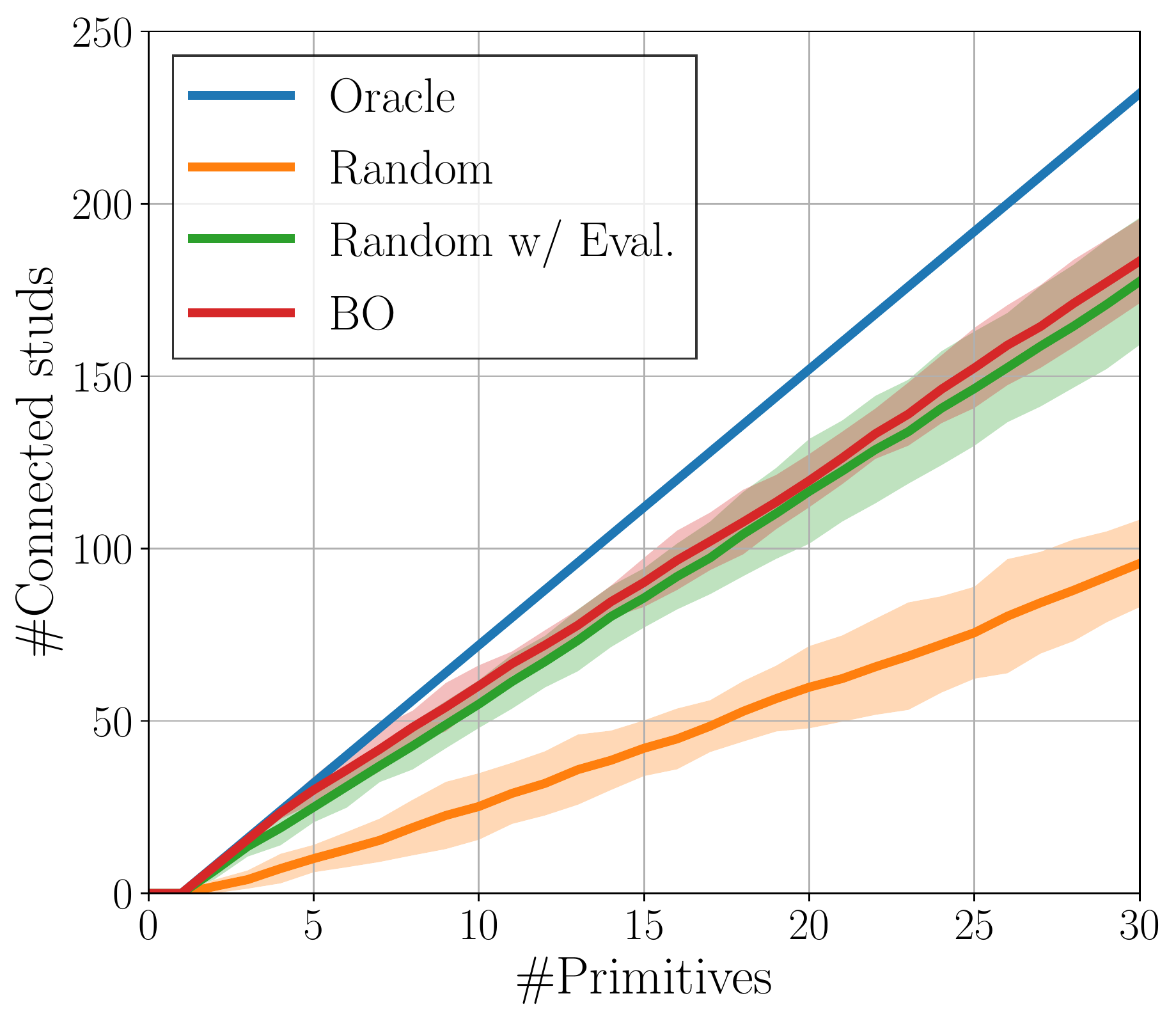}
        \label{fig:studs}
    }
    \caption{Quantitative results on maximizing four explicit evaluation functions (\ie, height, width, depth, and the number of connected studs).
    BO indicates our sequential assembly method.
    All the experiments are repeated ten times, 
    and their means and the 1.96 standard deviations are plotted.}
    \label{fig:ablation_functions}
\end{figure}

\begin{figure}[t]
    \centering
    \renewcommand{\thesubfigure}{(A-\alph{subfigure})}
    \setcounter{subfigure}{0}
    \subfigure[Parallel]
    {
        \includegraphics[width=0.12\textwidth]{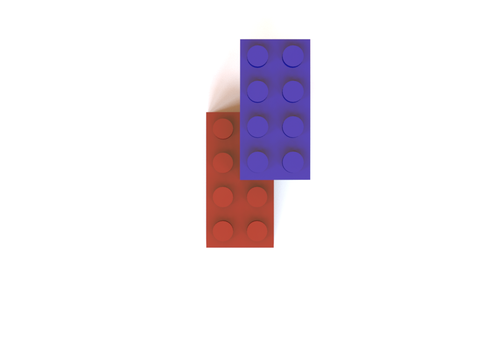}
        \label{fig:dataset_parallel_1}
    }
    \subfigure[Perpen.]
    {
        \includegraphics[width=0.12\textwidth]{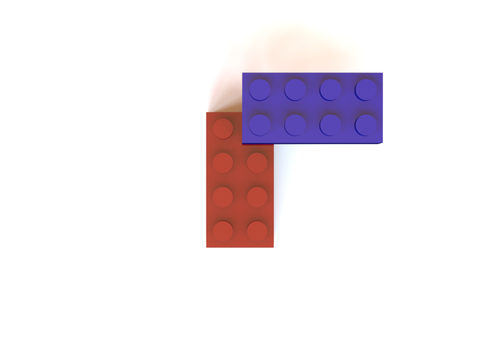}
        \label{fig:dataset_perpendicular_1}
    }
    \renewcommand{\thesubfigure}{(B-\alph{subfigure})}
    \setcounter{subfigure}{0}
    \subfigure[Bar]
    {
        \includegraphics[width=0.12\textwidth]{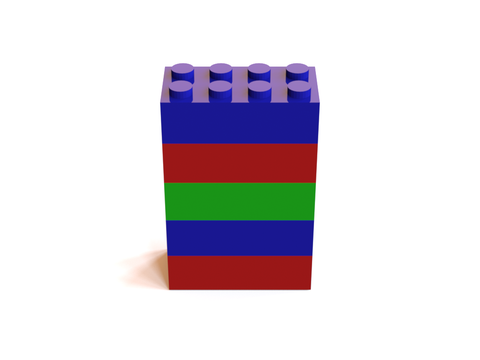}
        \label{fig:dataset_bar_1}
    }
    \subfigure[Line]
    {
        \includegraphics[width=0.12\textwidth]{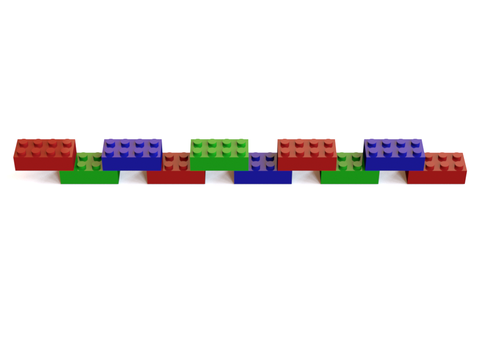}
        \label{fig:dataset_line_1}
    }
    \subfigure[Plate]
    {
        \includegraphics[width=0.12\textwidth]{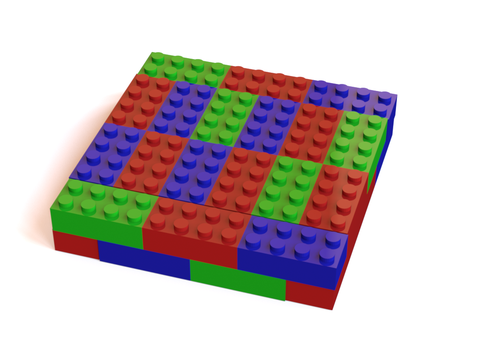}
        \label{fig:dataset_plate_1}
    }
    \subfigure[Wall]
    {
        \includegraphics[width=0.12\textwidth]{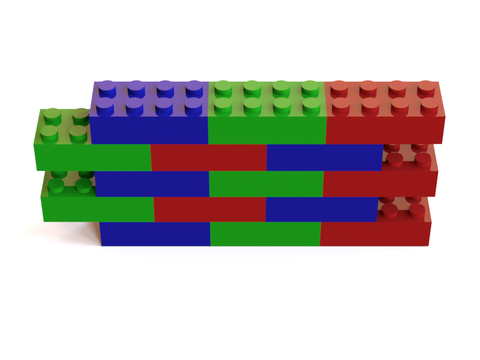}
        \label{fig:dataset_wall_1}
    }
    \subfigure[Cuboid]
    {
        \includegraphics[width=0.12\textwidth]{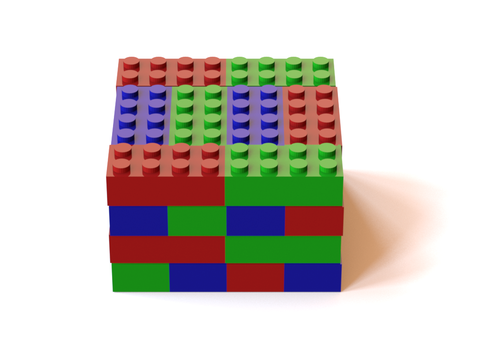}
        \label{fig:dataset_cuboid_1}
    }
    \subfigure[Pyramid]
    {
        \includegraphics[width=0.12\textwidth]{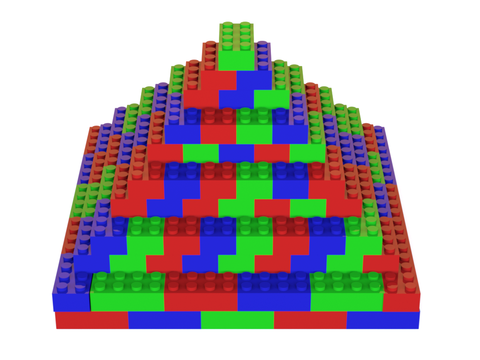}
        \label{fig:dataset_pyramid_1}
    }
    \renewcommand{\thesubfigure}{(C-\alph{subfigure})}
    \setcounter{subfigure}{0}
    \subfigure[Bench]
    {
        \includegraphics[width=0.12\textwidth]{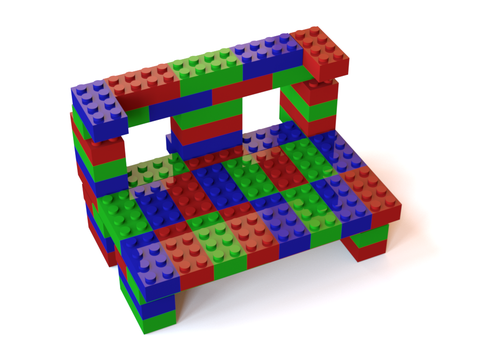}
        \label{fig:dataset_bench_1}
    }
    \subfigure[Sofa]
    {
        \includegraphics[width=0.12\textwidth]{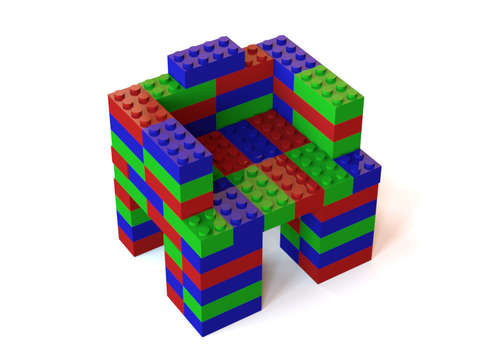}
        \label{fig:dataset_sofa_1}
    }
    \subfigure[Cup]
    {
        \includegraphics[width=0.12\textwidth]{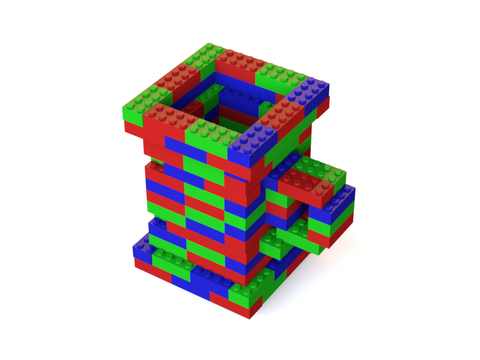}
        \label{fig:dataset_cup_1}
    }
    \subfigure[Hollow]
    {
        \includegraphics[width=0.12\textwidth]{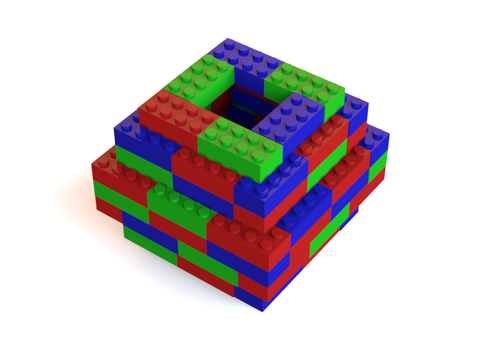}
        \label{fig:dataset_hollow_1}
    }
    \subfigure[Table]
    {
        \includegraphics[width=0.12\textwidth]{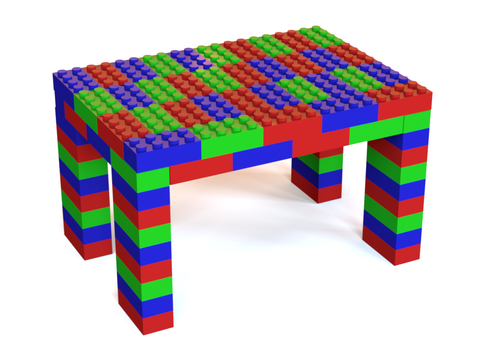}
        \label{fig:dataset_table_1}
    }
    \subfigure[Car]
    {
        \includegraphics[width=0.12\textwidth]{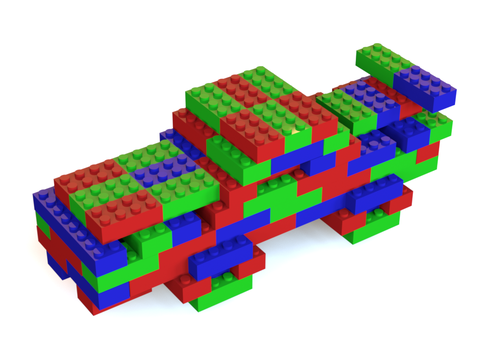}
        \label{fig:dataset_car_1}
    }
    \caption{Selected examples from our combinatorial 3D shape dataset.}
    \label{fig:dataset_all}
\end{figure}

\paragraph{Rolling Back the Primitives Previously Assembled}
Our method might be able to provide a sub-optimal sequence, 
because the Bayesian optimization module, which always guarantees to find a global solution~\citep{BrochuE2010arxiv,SrinivasN2010icml}, 
accumulates local (but possibly global) optima in a row.
Hence, our method includes a rollback step, as shown in Line~7 to Line~10 in \algref{alg:combination}.
Given the number of rollback steps $\tau$ and a threshold for rolling back $\alpha$, 
if the condition written below is satisfied:
$t \geq \tau$ and 
$\sum_{k = 0}^{\tau - 1} \left( \max_{\bp_{m + t - k}} y_{\occ, t - k} \right) - y_{\occ, t - k} < \alpha$,
$t$-primitive combination is rolled back to $(t - \tau)$-primitive combination.
To avoid rolling back in the same combination infinitely, 
we prevent placing the same position if the rollback is activated and 
we skip the rollback if the same assembling step is repeated for five times.

\section{Experimental Results\label{sec:exp}}

\paragraph{Combinatorial 3D Shape Generation via Sequential Assembly}
We sequentially generate a combinatorial 3D shape 
by optimizing \eqqref{eqn:scores_occ_sta}, 
given the occupiability with a desired shape, 
as described in the main method section.
Our assembly results shown in \figref{fig:teaser} and \figref{fig:assembly_cup_bench} 
demonstrate that our method can effectively find a feasible combination, 
by considering our evaluation functions.
To accommodate a page limit, we provide additional experimental results in the appendices.

\paragraph{Explicit Evaluation Functions for Bayesian Optimization Module}
To confirm the validity of our Bayesian optimization module where an evident evaluation function is given.
For example, the height of combination can express 
the current status of the given combination as an evaluation score:
\begin{equation}
    f \left(\bp_{n + 1}; \{ \bp_i \}_{i = 1}^n \right) = \max_{(x_1, x_2, x_3, d) \in \{\bp_i\}_{i = 1}^{n + 1}} x_3 + 1.
    \label{eqn:height}
\end{equation}
Similar to \eqqref{eqn:height}, we define three more specific functions: width, depth, and the number of connected studs.
Likewise, these BO modules attempt to maximize the evaluation functions we define.
For these experiments, we use three baselines, which are described in the appendices.
As shown in \figref{fig:ablation_functions}, our method outperforms other methods and achieves the results closest to the oracles of four experiments.

\section{Combinatorial 3D Shape Dataset\label{sec:dataset}}

We construct a combinatorial 3D shape dataset, composed of 406 instances of 14 classes.
Specifically, each object in our dataset is considered equivalent to a sequence of primitive placement.
For this reason, compared to other 3D object datasets~\citep{ChangAX2015arxiv,XiangY2016eccv}, 
our proposed dataset contains an assembling sequence of unit primitives.
It implies that we can quickly obtain a sequential generation process 
that is a human assembling mechanism.
Furthermore, we can sample valid random sequences from a given combinatorial shape 
after validating the sampled sequences.
To sum up, the characteristics of our combinatorial 3D shape dataset are 
\begin{enumerate}[label=(\roman*)]
    \item combinatorial: Duplicates of unit primitive is repeatedly connected;
    \item sequential: Allowable connections between primitives are sequentially added;
    \item decomposable: By the combinatorial property, parts of combination can be sampled if they are valid in terms of the contact and overlap conditions;
    \item manipulable: New primitive is addable or the existing primitives are removable.
\end{enumerate}

Our 3D shape dataset is implemented to satisfy the above properties, 
supporting sequential assembly, combination validation, possible position listing, and part sampling.

As shown in \figref{fig:dataset_all}, we select 14 classes: 
parallel, perpendicular, bar, line, plate, wall, cuboid, square pyramid, bench, sofa, cup, hollow, table, and car.
Parallel that implies the directions of two primitives are same, 
and perpendicular that implies the directions of two primitives are different classes are own connection types of $\twobyfour$-sized primitives with studs and cavities (denoted as group A).
Bar, plate, cuboid, wall, and square pyramid classes are taken into account as default components to assemble sophisticated shapes (denoted as group B).
The other classes are abstractly thought of as the combination of those default components (denoted as group C).
More diverse examples and the statistics of our dataset can be found in the appendices.

\section{Conclusion\label{sec:conclusion}}

We propose a sequential assembly method for a combinatorial 3D generation problem.
It can generate a combinatorial shape, considering sample efficiency 
that is guided by Bayesian optimization.
The evaluation function based on global shape guidance and stability demonstrates 
that our method generates 3D shapes composed of unit primitives.
Besides, we create a new dataset for combinatorial 3D models.
This dataset allows us to generate 3D shapes sequentially.

\begin{ack}
This work was supported by Institute for Information \& communications Technology Promotion (IITP) grant funded by the Korea government (MSIP) (No.2019-0-01906, Artificial Intelligence Graduate School Program (POSTECH)).
\end{ack}

\bibliographystyle{abbrvnat}
\bibliography{sequentialassembly}

\begin{appendices}
\section*{Appendices}

We describe the details which are omitted from the main article.
First, we show the connection types between two $\twobyfour$-sized primitives.
Then, we describe the experimental setups 
and visualize some examples of our combinatorial 3D shape dataset.

\section{Connection Types between Two Two-by-Four Primitives}

\begin{figure}[ht]
    \centering
        \includegraphics[width=0.11\textwidth]{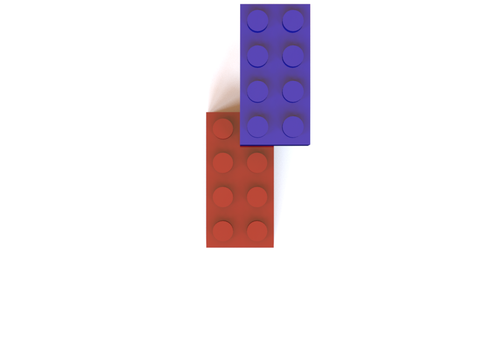}
        \includegraphics[width=0.11\textwidth]{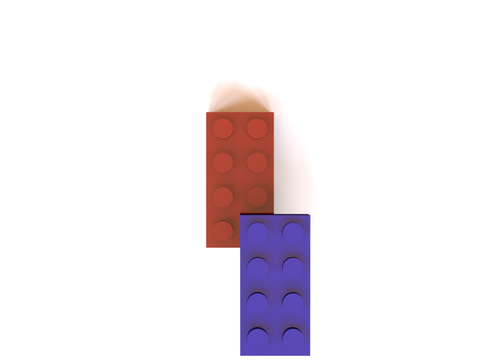}
        \includegraphics[width=0.11\textwidth]{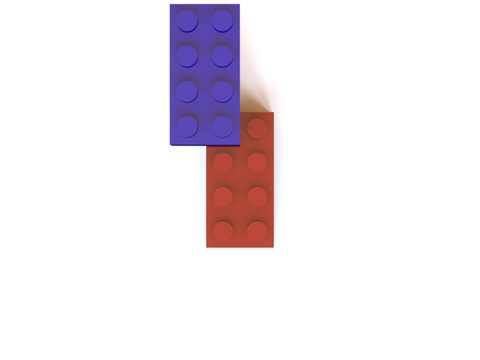}
        \includegraphics[width=0.11\textwidth]{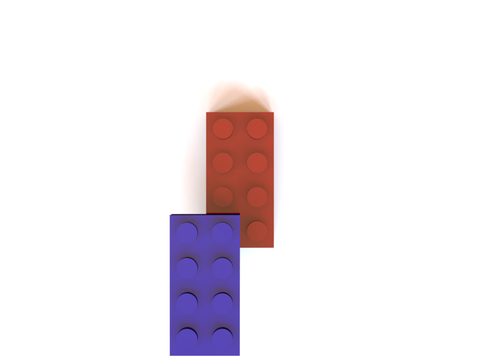}
        \includegraphics[width=0.11\textwidth]{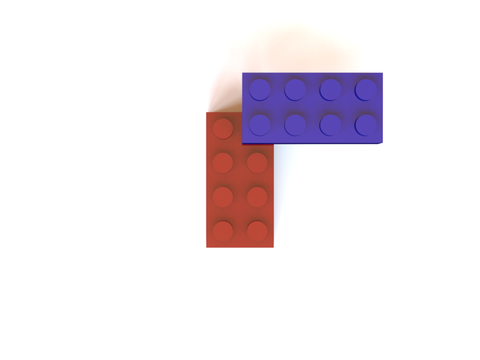}
        \includegraphics[width=0.11\textwidth]{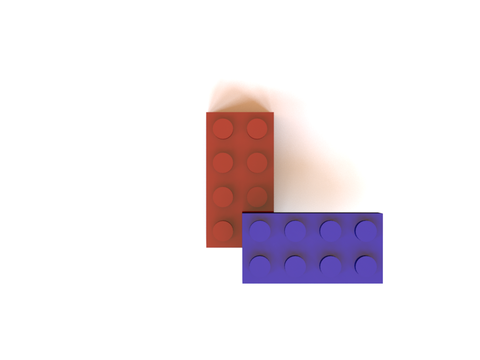}
        \includegraphics[width=0.11\textwidth]{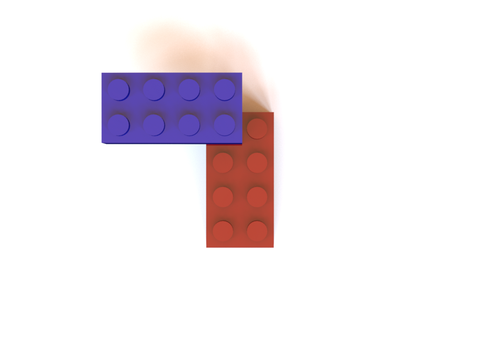}
        \includegraphics[width=0.11\textwidth]{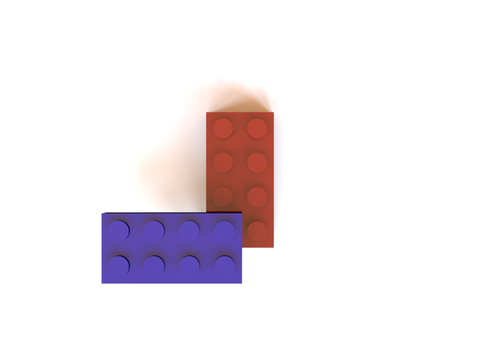}
        \includegraphics[width=0.11\textwidth]{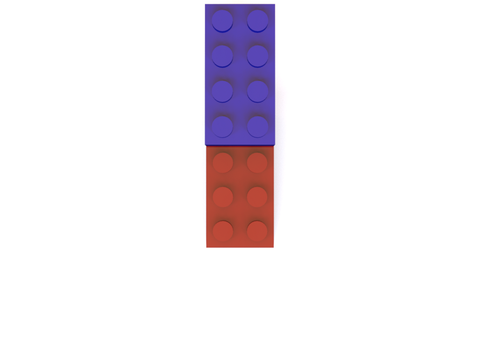}
        \includegraphics[width=0.11\textwidth]{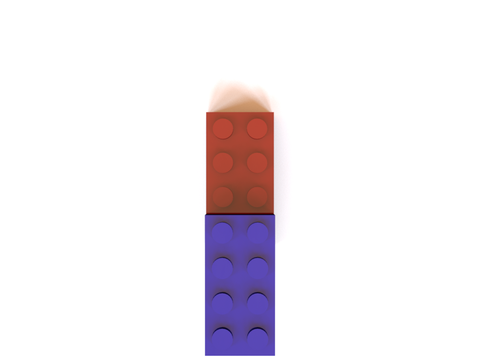}
        \includegraphics[width=0.11\textwidth]{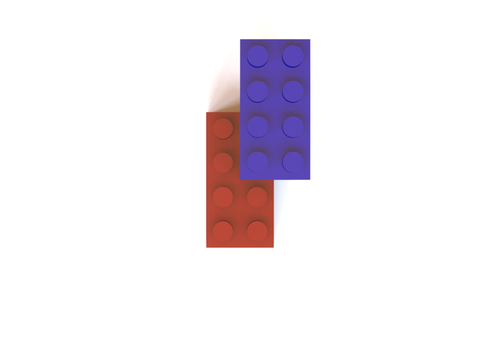}
        \includegraphics[width=0.11\textwidth]{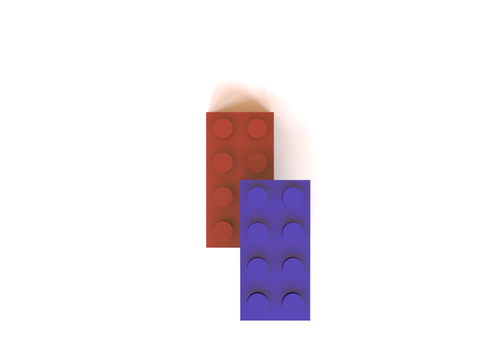}
        \includegraphics[width=0.11\textwidth]{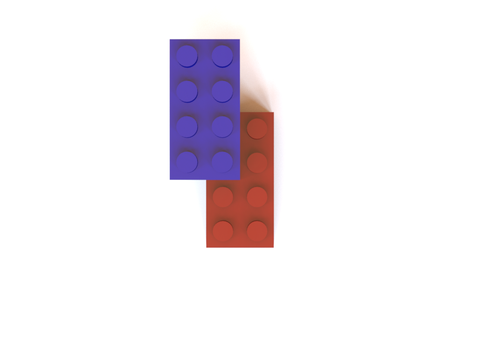}
        \includegraphics[width=0.11\textwidth]{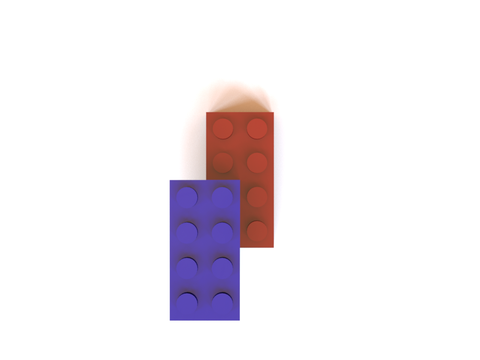}
        \includegraphics[width=0.11\textwidth]{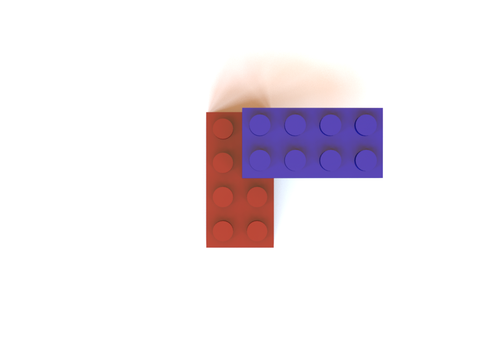}
        \includegraphics[width=0.11\textwidth]{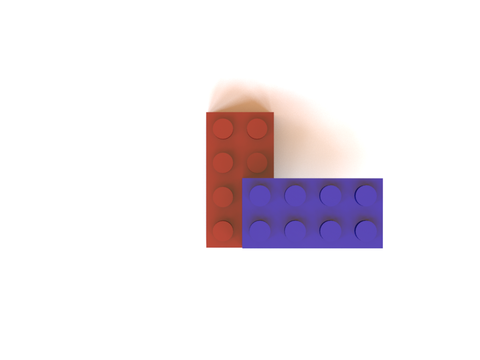}
        \includegraphics[width=0.11\textwidth]{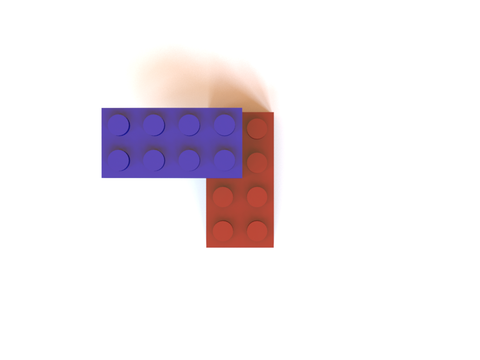}
        \includegraphics[width=0.11\textwidth]{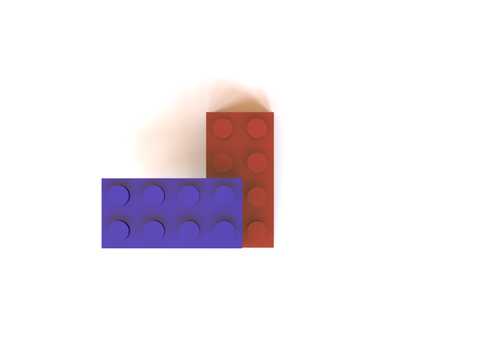}
        \includegraphics[width=0.11\textwidth]{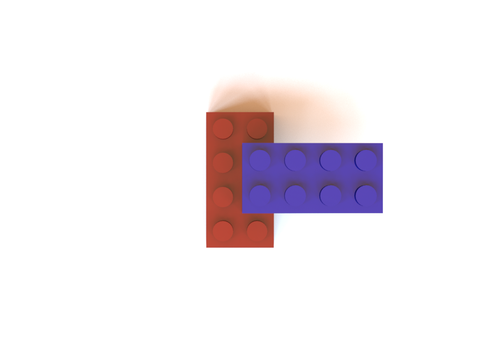}
        \includegraphics[width=0.11\textwidth]{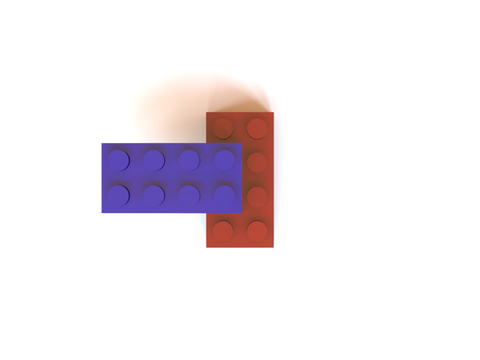}
        \includegraphics[width=0.11\textwidth]{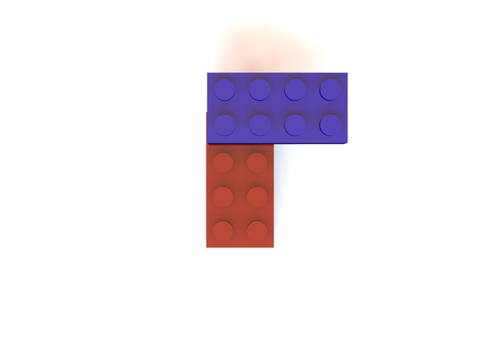}
        \includegraphics[width=0.11\textwidth]{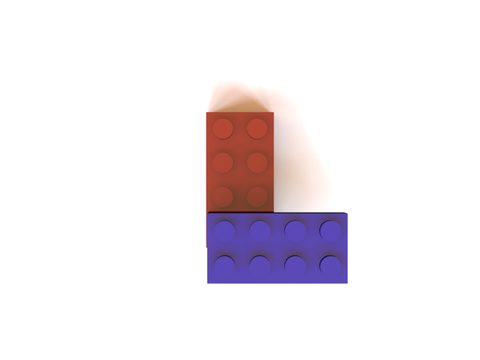}
        \includegraphics[width=0.11\textwidth]{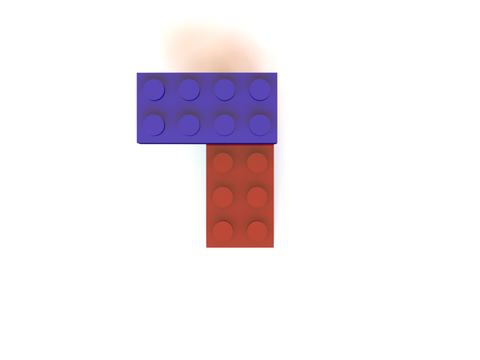}
        \includegraphics[width=0.11\textwidth]{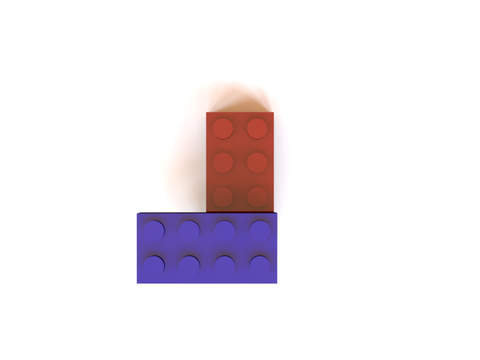}
        \includegraphics[width=0.11\textwidth]{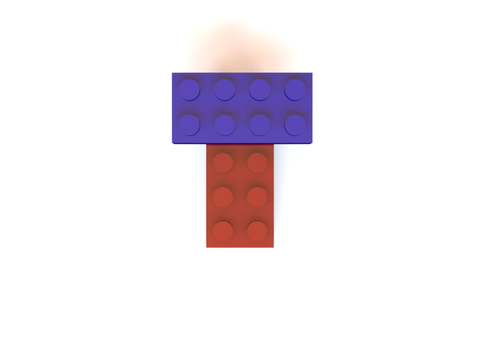}
        \includegraphics[width=0.11\textwidth]{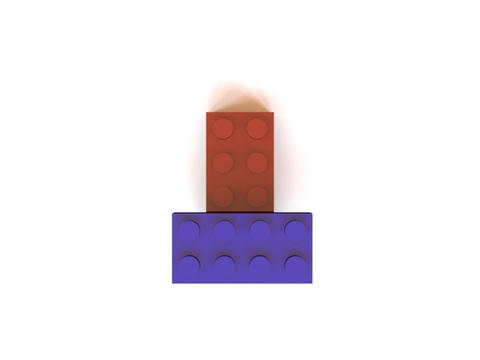}
        \includegraphics[width=0.11\textwidth]{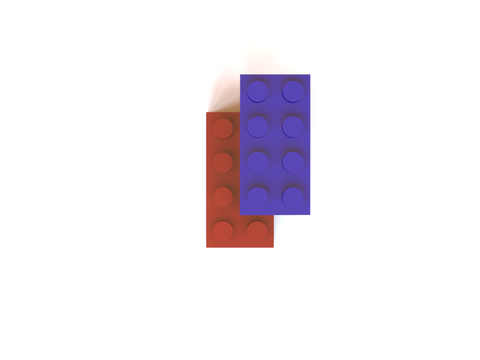}
        \includegraphics[width=0.11\textwidth]{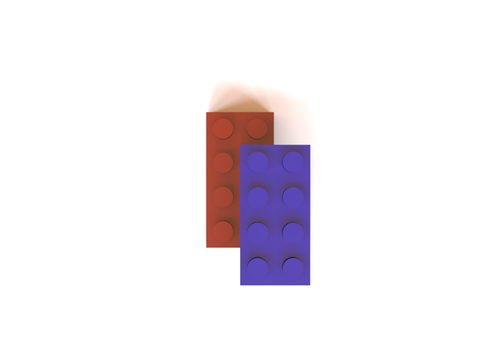}
        \includegraphics[width=0.11\textwidth]{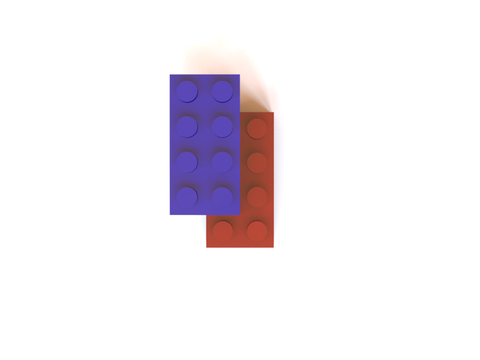}
        \includegraphics[width=0.11\textwidth]{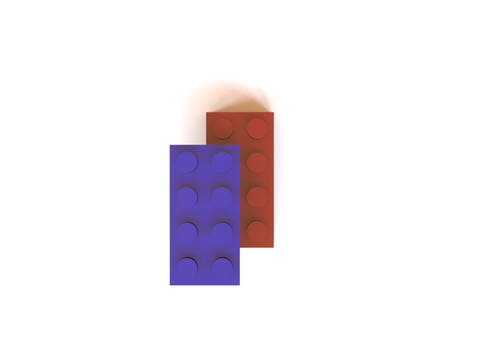}
        \includegraphics[width=0.11\textwidth]{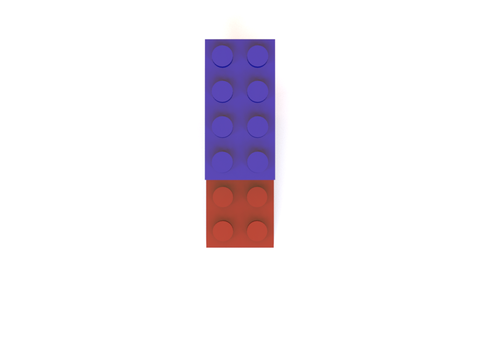}
        \includegraphics[width=0.11\textwidth]{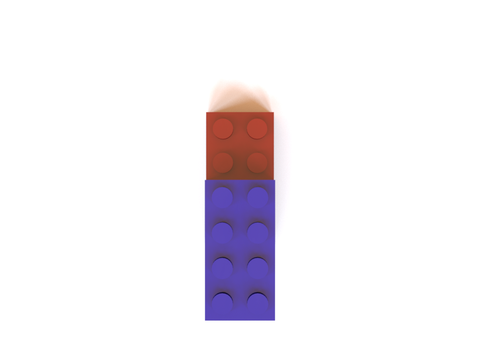}
        \includegraphics[width=0.11\textwidth]{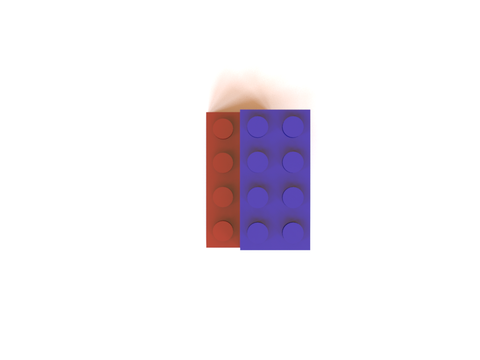}
        \includegraphics[width=0.11\textwidth]{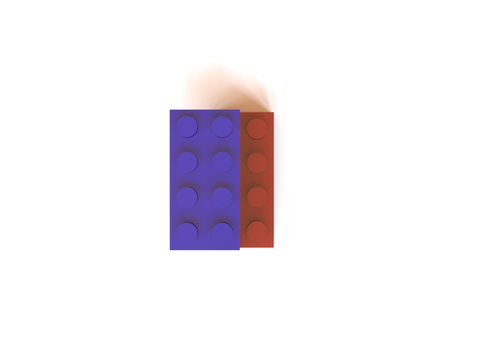}
        \includegraphics[width=0.11\textwidth]{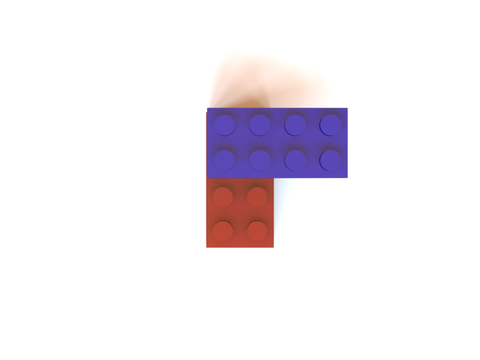}
        \includegraphics[width=0.11\textwidth]{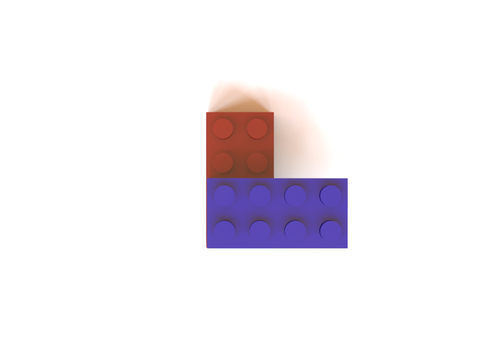}
        \includegraphics[width=0.11\textwidth]{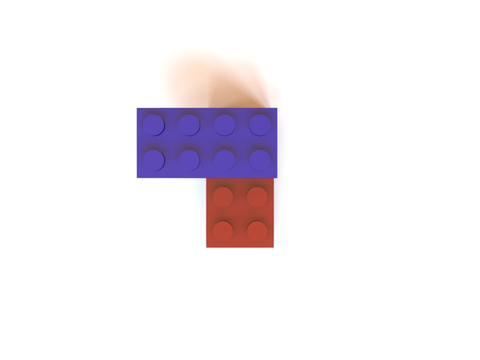}
        \includegraphics[width=0.11\textwidth]{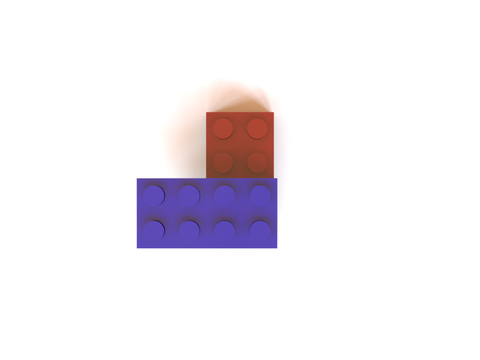}
        \includegraphics[width=0.11\textwidth]{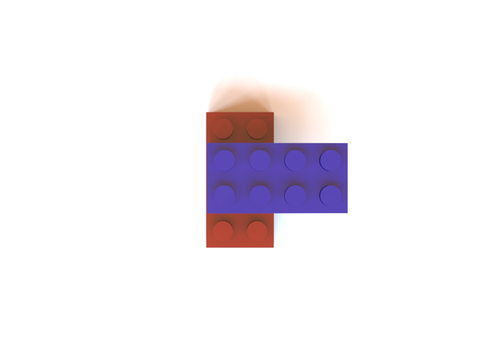}
        \includegraphics[width=0.11\textwidth]{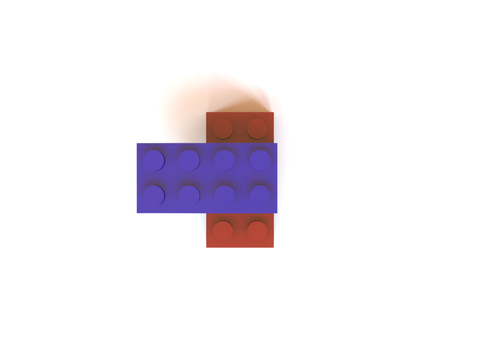}
        \includegraphics[width=0.11\textwidth]{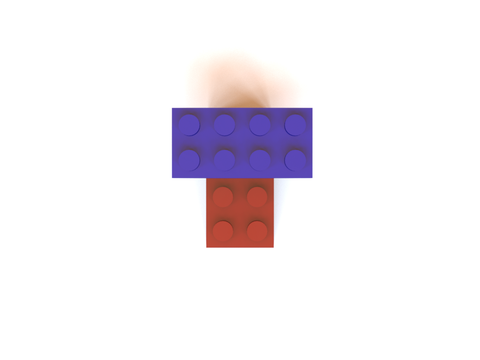}
        \includegraphics[width=0.11\textwidth]{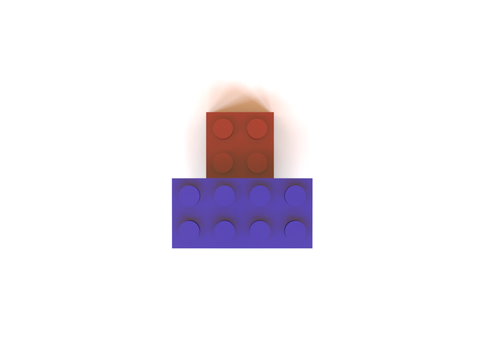}
        \includegraphics[width=0.11\textwidth]{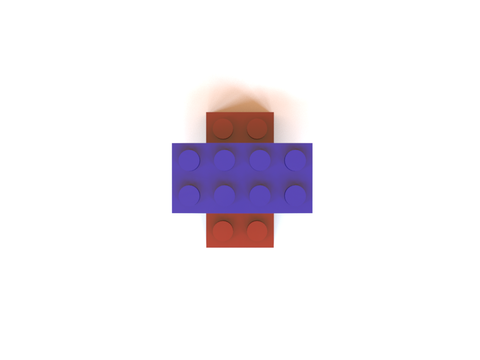}
        \includegraphics[width=0.11\textwidth]{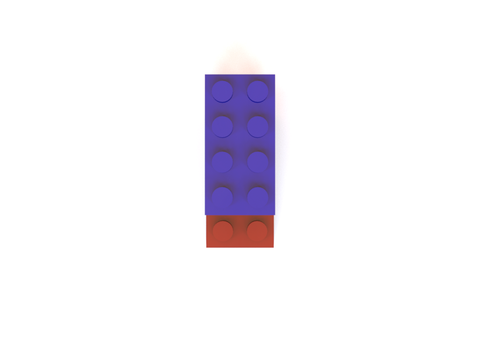}
        \includegraphics[width=0.11\textwidth]{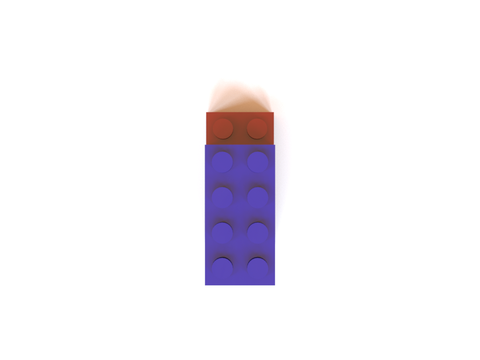}
        \includegraphics[width=0.11\textwidth]{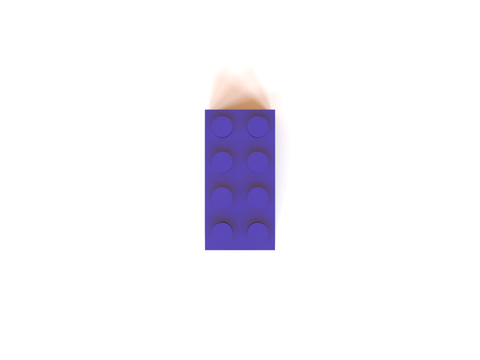}
    \caption{46 connection types between two $\twobyfour$-sized primitives.}
    \label{fig:suppl_connection_types}
\end{figure}

There exist 46 connection types between two $\twobyfour$ LEGO brick-shaped primitives 
where upper and lower primitives are fixed, 
as shown in \figref{fig:suppl_connection_types}.
They comprise group A of our combinatorial dataset.

\section{Experimental Setups}

Gaussian process regression with Mat\'ern 5/2 kernels~\citep{RasmussenCE2006book} 
and Gaussian process upper confidence bound (GP-UCB)~\citep{SrinivasN2010icml} 
are used as a surrogate model and an acquisition function, respectively.
The hyperparameters of kernels for the regression models are optimized 
by marginal likelihood maximization with BFGS algorithm.
Similar to the setting in \citep{SrinivasN2010icml}, the trade-off hyperparameter 
of GP-UCB monotonically increases over iterations.
Rather than setting a specific number of samples $\zeta$ for Line~4 of \algref{alg:select}, 
we sample as many candidates of maximizer as possible within the given time budget.
We set the time budget as 1 second in most of the cases.
Unless otherwise specified, $v$ and $q$ in \algref{alg:select} are set to 10 and 20, respectively.
In all the experiments, an initial primitive combination in \algref{alg:combination} 
is given as $\{([0, 0, 0], 0)\}$, to assemble from scratch.
Bayesian optimization modules are implemented with \citep{KimJ2017bayeso}.
For multi-objective Bayesian optimization, $\lambda_{\occ}$ is sampled from 
uniform distribution $\calU(0.8, 0.9)$ 
and $\lambda_{\sta}$ is sampled from uniform distribution $\calU(0.0, 0.1)$.

\begin{figure}[t]
    \centering
    \subfigure[Placing]
    {
        \includegraphics[width=0.32\textwidth]{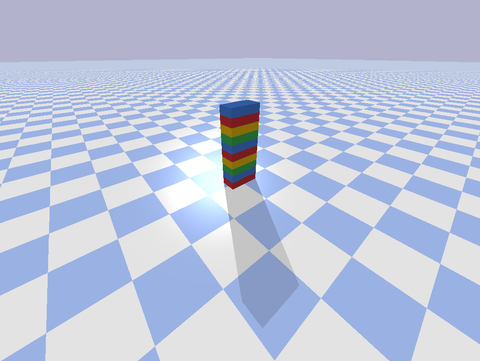}
        \label{fig:placing}
    }
    \subfigure[Applying forces]
    {
        \includegraphics[width=0.3\textwidth]{pybullet_3.pdf}
        \label{fig:applying}
    }
    \subfigure[Measuring stability]
    {
        \includegraphics[width=0.32\textwidth]{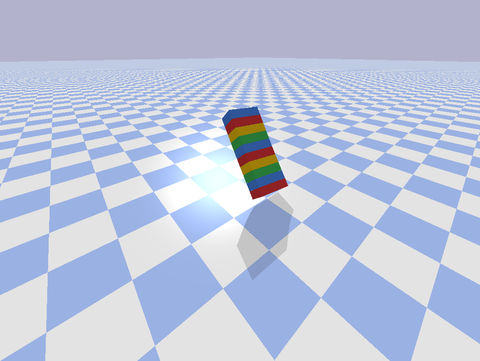}
        \label{fig:measuring}
    }
    \caption{Stability simulation results with \texttt{PyBullet}.}
    \label{fig:suppl_pybullet}
\end{figure}

To measure the stability, we use the physics engine simulator~\citep{CoumansE2019pybullet}
as shown in \figref{fig:suppl_pybullet}.
We first place a combination we would like to measure the stability.
And then, we alternately apply external forces to one of four directions (see \figref{fig:applying}) 
at the beginning of the simulation (\ie, 200 time steps) 
where gravity is existed.
After applying the forces, we measure time steps to be stable in terms of the position of given combination.

Open3D framework~\citep{ZhouQ2018arxiv} and Mitsuba renderer~\citep{JakobW2010mitsuba} are employed 
to deal with 3D objects and visualize the primitive assembly results.
For attractive visualization, we randomly pick the color of primitives from red, blue, and green.

For the experiments for explicit evaluation functions, we compare our method to three baselines:
\begin{enumerate}[label=(\roman*)]
    \item Oracle: It is the best achievable result;
    \item Random: This baseline is a fully-randomized result.
    One of the primitives possible to assemble is uniformly selected at every assembling step;
    \item Random with evaluations: It is a result by a greedy method.
    The best primitive is chosen at every step after evaluating the primitives uniformly sampled.
    The number of the sampled primitives is set to $q$ to compare with our method fairly.    
\end{enumerate}

\section{Combinatorial 3D Shape Dataset}

\begin{figure}[ht]
    \centering
        \includegraphics[width=0.19\textwidth]{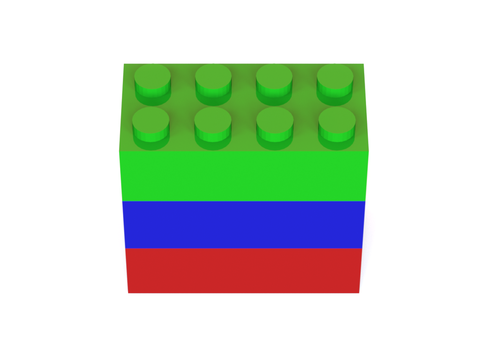}
        \includegraphics[width=0.19\textwidth]{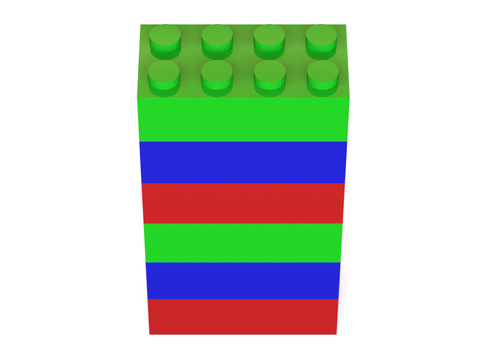}
        \includegraphics[width=0.19\textwidth]{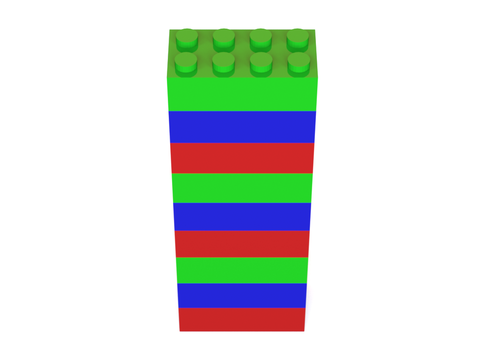}
        \includegraphics[width=0.19\textwidth]{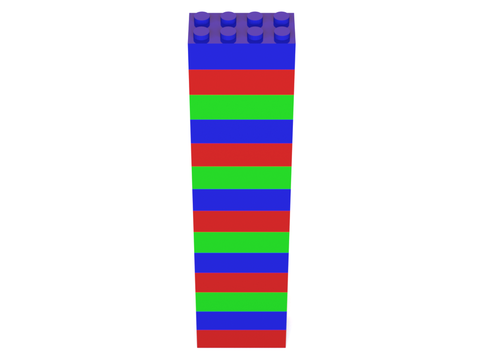}
        \includegraphics[width=0.19\textwidth]{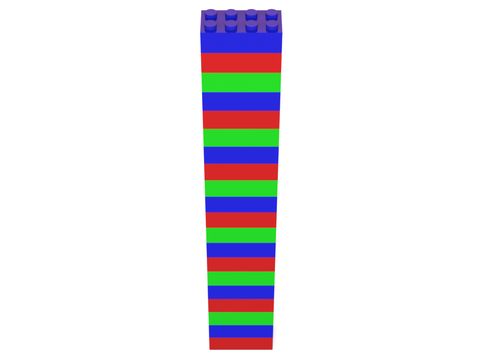}
    \caption{Examples of \emph{Bar} class.}
    \label{fig:suppl_dataset_bar}
\end{figure}

\begin{figure}[ht]
    \centering
        \includegraphics[width=0.19\textwidth]{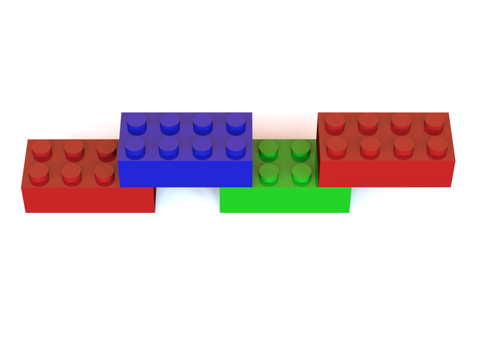}
        \includegraphics[width=0.19\textwidth]{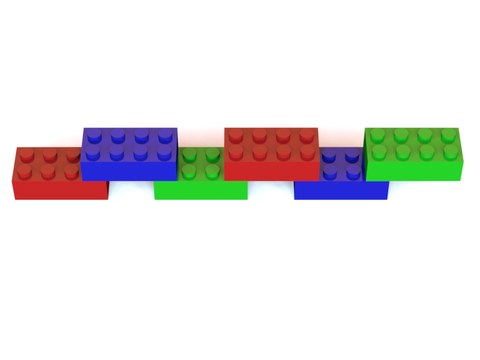}
        \includegraphics[width=0.19\textwidth]{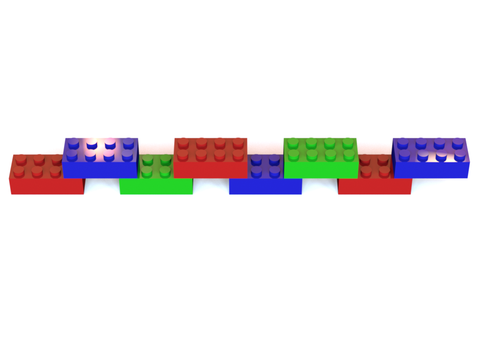}
        \includegraphics[width=0.19\textwidth]{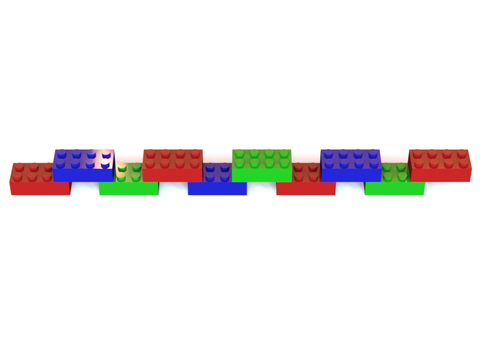}
        \includegraphics[width=0.19\textwidth]{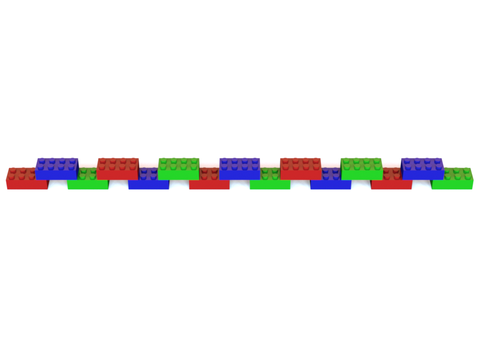}
    \caption{Examples of \emph{Line} class.}
    \label{fig:suppl_dataset_line}
\end{figure}

\begin{figure}[ht]
    \centering
        \includegraphics[width=0.19\textwidth]{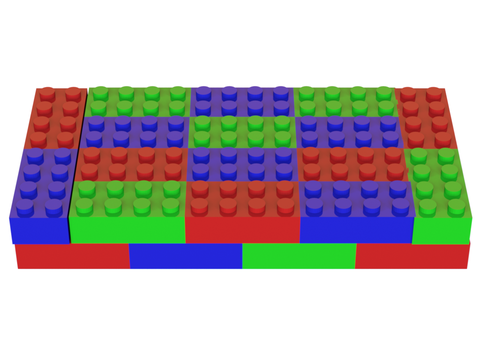}
        \includegraphics[width=0.19\textwidth]{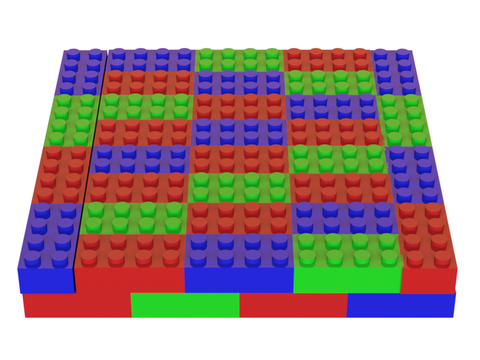}
        \includegraphics[width=0.19\textwidth]{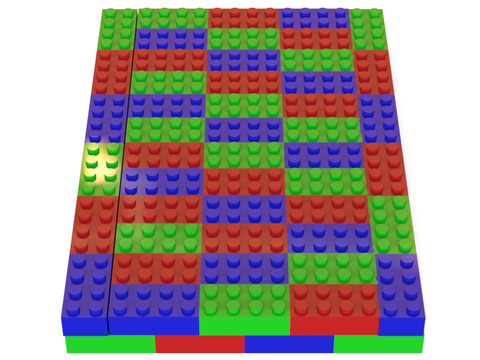}
        \includegraphics[width=0.19\textwidth]{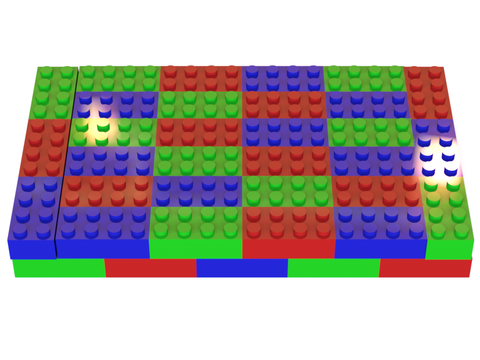}
        \includegraphics[width=0.19\textwidth]{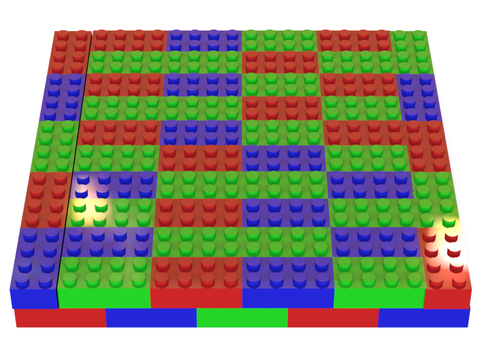}
    \caption{Examples of \emph{Plate} class.}
    \label{fig:suppl_dataset_plate}
\end{figure}

\begin{figure}[ht]
    \centering
        \includegraphics[width=0.19\textwidth]{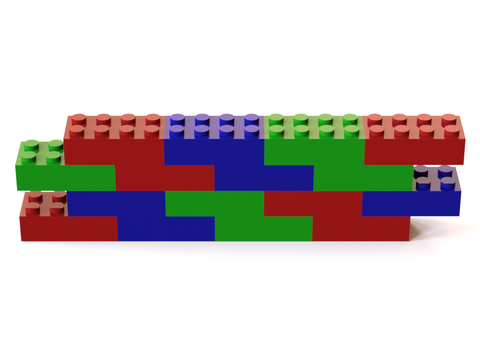}
        \includegraphics[width=0.19\textwidth]{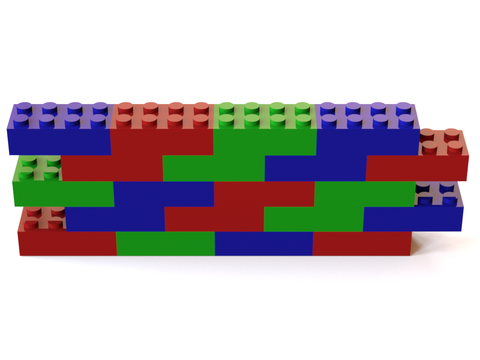}
        \includegraphics[width=0.19\textwidth]{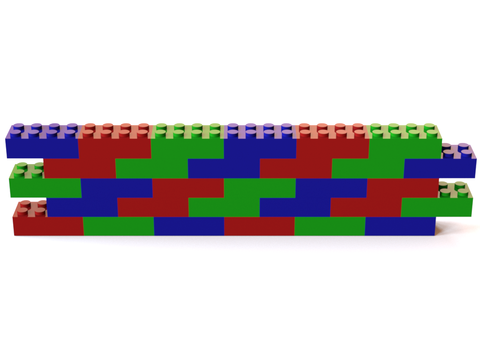}
        \includegraphics[width=0.19\textwidth]{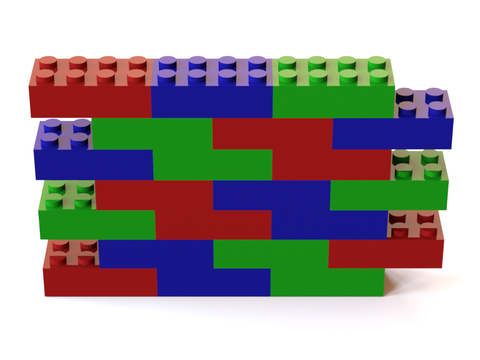}
        \includegraphics[width=0.19\textwidth]{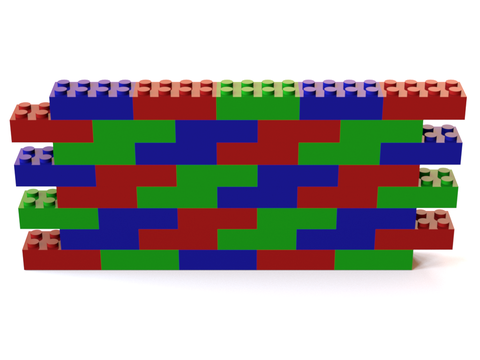}
    \caption{Examples of \emph{Wall} class.}
    \label{fig:suppl_dataset_wall}
\end{figure}

\begin{figure}[ht]
    \centering
        \includegraphics[width=0.19\textwidth]{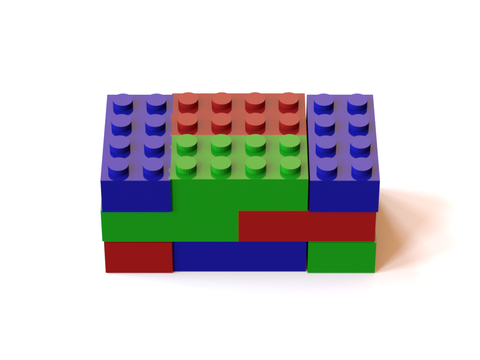}
        \includegraphics[width=0.19\textwidth]{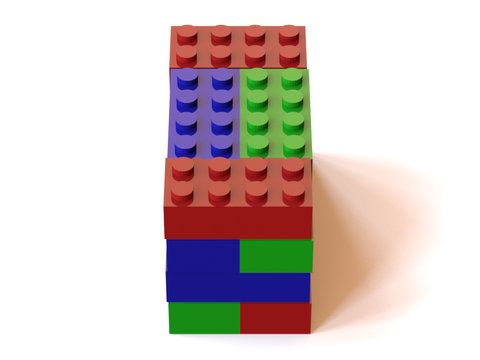}
        \includegraphics[width=0.19\textwidth]{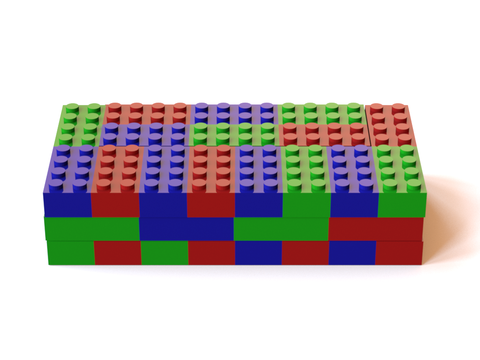}
        \includegraphics[width=0.19\textwidth]{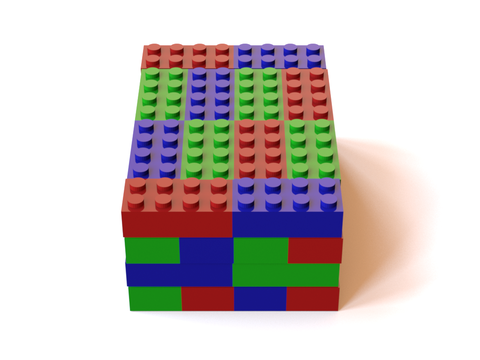}
        \includegraphics[width=0.19\textwidth]{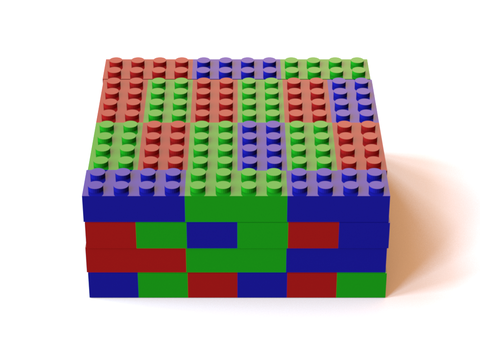}
    \caption{Examples of \emph{Cuboid} class.}
    \label{fig:suppl_dataset_cuboid}
\end{figure}

\begin{figure}[ht]
    \centering
        \includegraphics[width=0.19\textwidth]{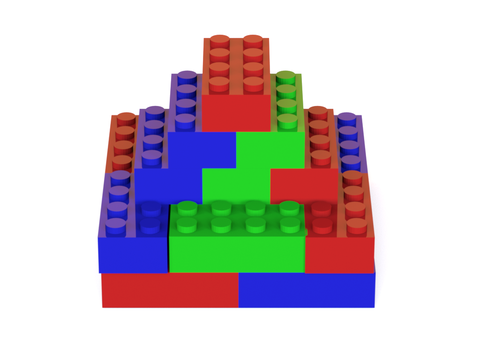}
        \includegraphics[width=0.19\textwidth]{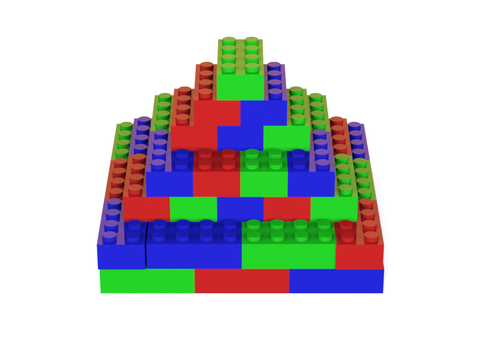}
        \includegraphics[width=0.19\textwidth]{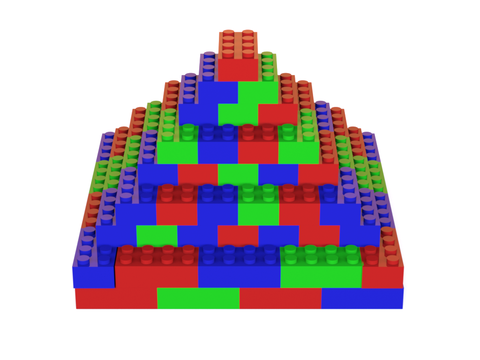}
        \includegraphics[width=0.19\textwidth]{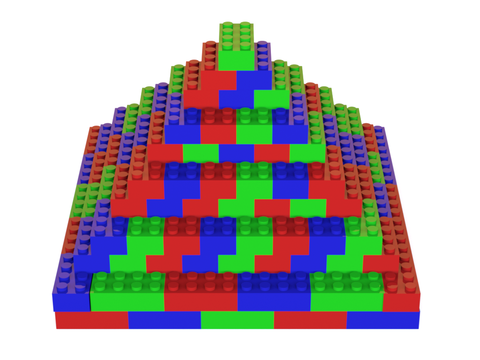}
        \includegraphics[width=0.19\textwidth]{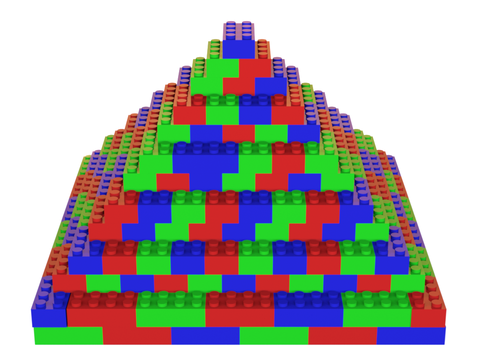}
    \caption{Examples of \emph{Square Pyramid} class.}
    \label{fig:suppl_dataset_pyramid}
\end{figure}

\begin{figure}[ht]
    \centering
        \includegraphics[width=0.19\textwidth]{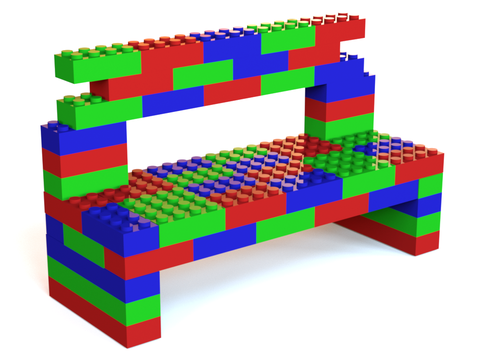}
        \includegraphics[width=0.19\textwidth]{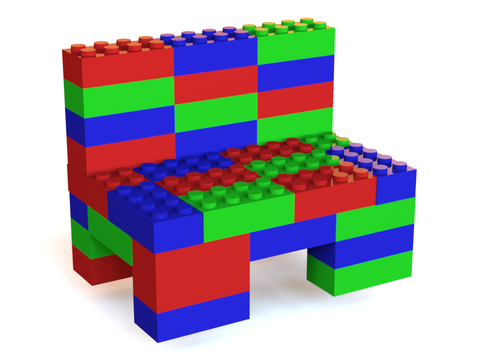}
        \includegraphics[width=0.19\textwidth]{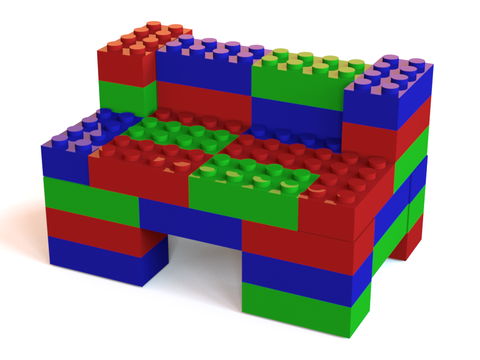}
        \includegraphics[width=0.19\textwidth]{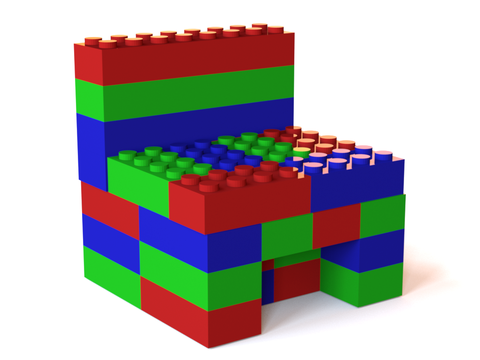}
        \includegraphics[width=0.19\textwidth]{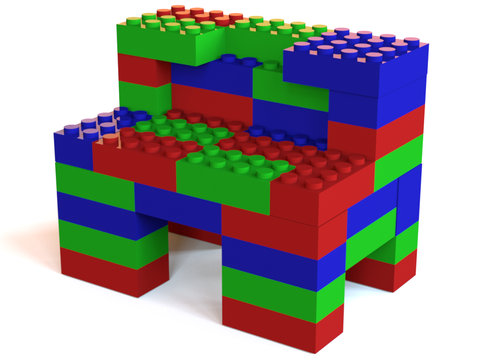}
    \caption{Examples of \emph{Bench} class.}
    \label{fig:suppl_dataset_bench}
\end{figure}

\begin{figure}[ht]
    \centering
        \includegraphics[width=0.19\textwidth]{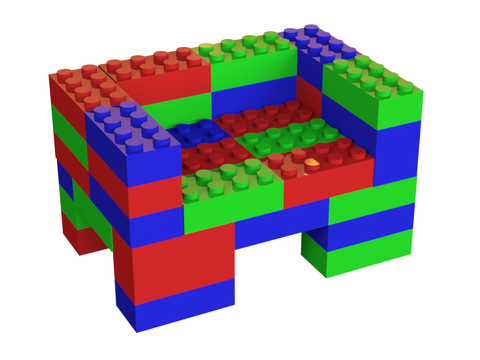}
        \includegraphics[width=0.19\textwidth]{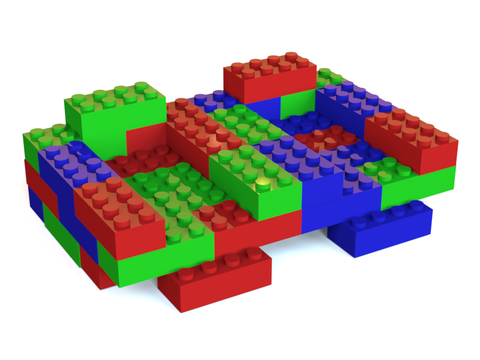}
        \includegraphics[width=0.19\textwidth]{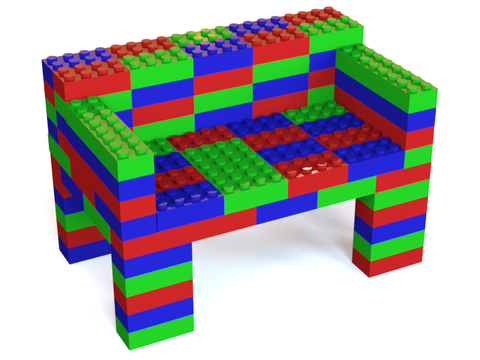}
        \includegraphics[width=0.19\textwidth]{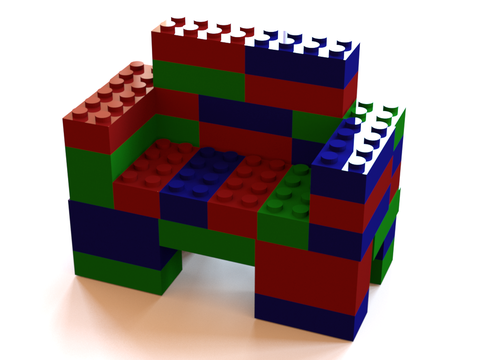}
        \includegraphics[width=0.19\textwidth]{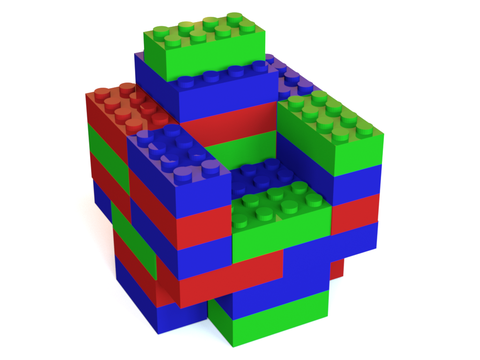}
    \caption{Examples of \emph{Sofa} class.}
    \label{fig:suppl_dataset_sofa}
\end{figure}

\begin{figure}[ht]
    \centering
        \includegraphics[width=0.19\textwidth]{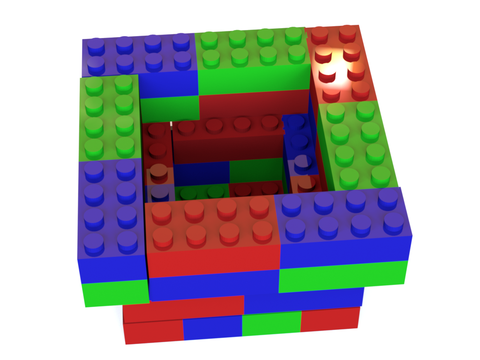}
        \includegraphics[width=0.19\textwidth]{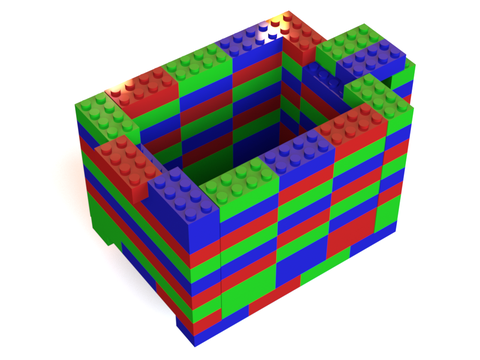}
        \includegraphics[width=0.19\textwidth]{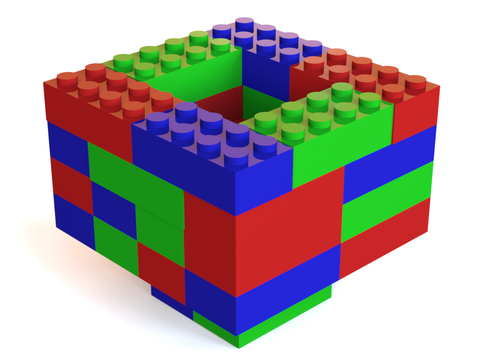}
        \includegraphics[width=0.19\textwidth]{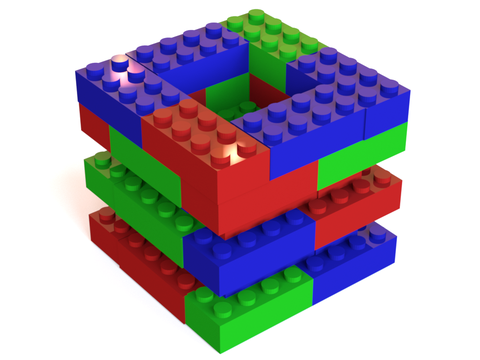}
        \includegraphics[width=0.19\textwidth]{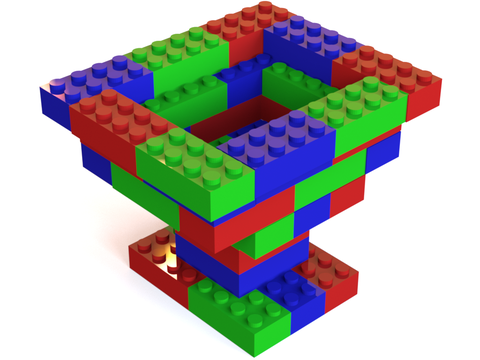}
    \caption{Examples of \emph{Cup} class.}
    \label{fig:suppl_dataset_cup}
\end{figure}

\begin{figure}[ht]
    \centering
        \includegraphics[width=0.19\textwidth]{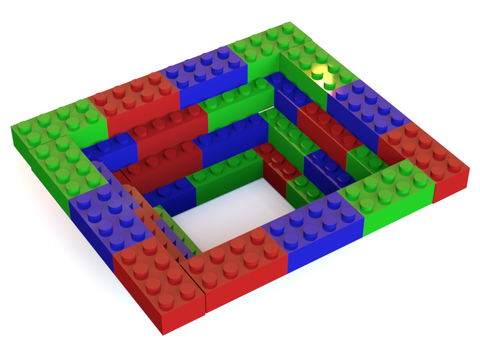}
        \includegraphics[width=0.19\textwidth]{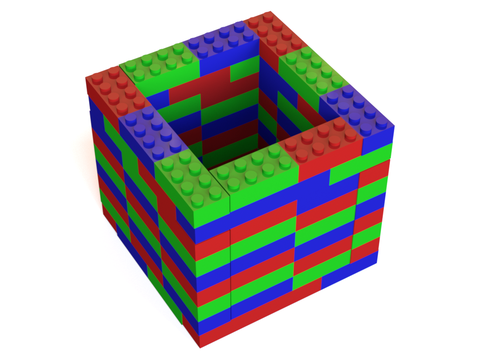}
        \includegraphics[width=0.19\textwidth]{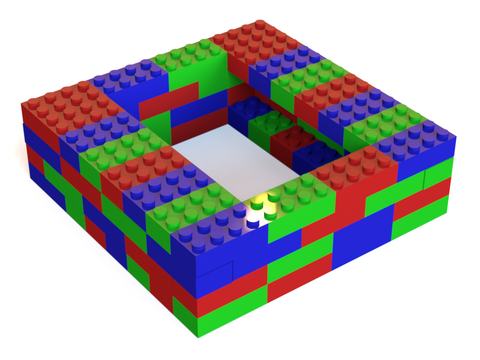}
        \includegraphics[width=0.19\textwidth]{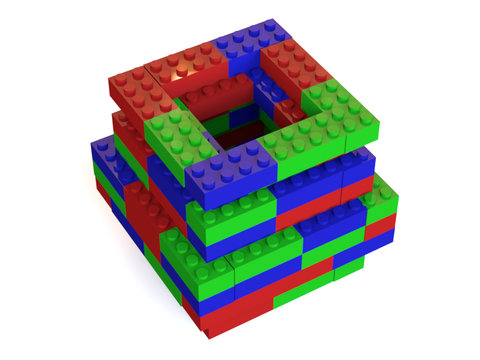}
        \includegraphics[width=0.19\textwidth]{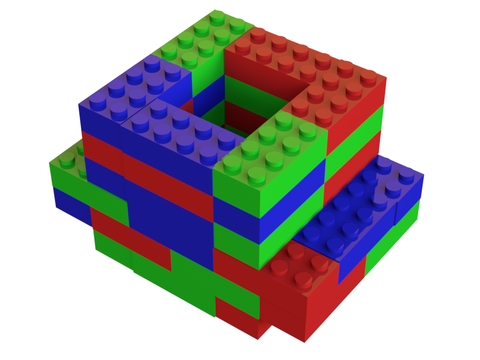}
    \caption{Examples of \emph{Hollow} class.}
    \label{fig:suppl_dataset_hollow}
\end{figure}

\begin{figure}[ht]
    \centering
        \includegraphics[width=0.19\textwidth]{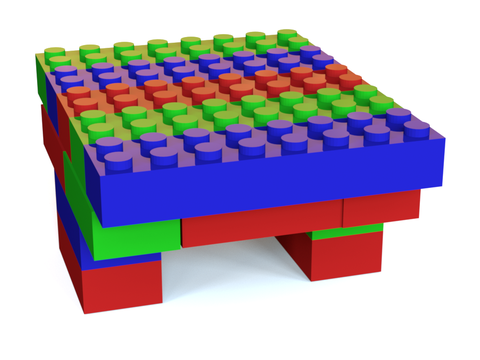}
        \includegraphics[width=0.19\textwidth]{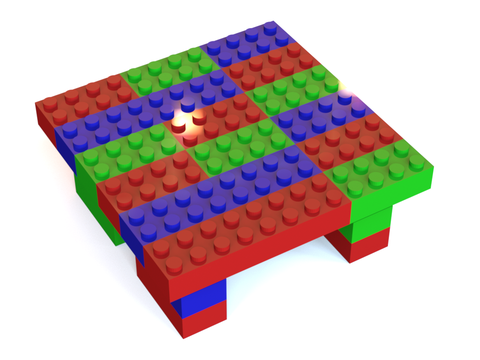}
        \includegraphics[width=0.19\textwidth]{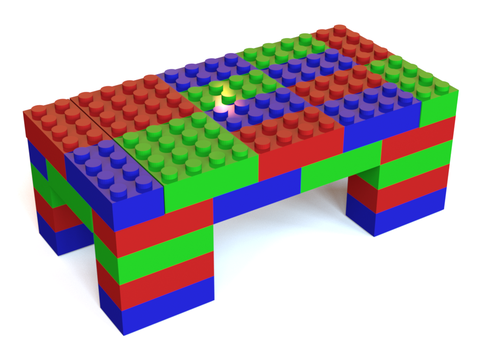}
        \includegraphics[width=0.19\textwidth]{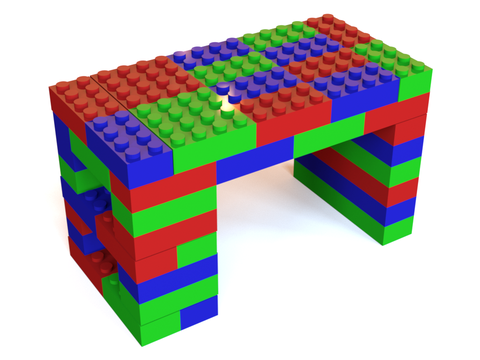}
        \includegraphics[width=0.19\textwidth]{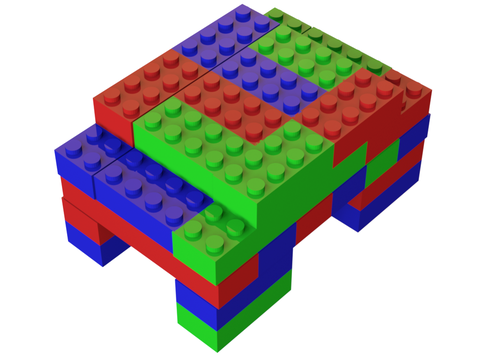}
    \caption{Examples of \emph{Table} class.}
    \label{fig:suppl_dataset_table}
\end{figure}

\begin{figure}[ht]
    \centering
        \includegraphics[width=0.19\textwidth]{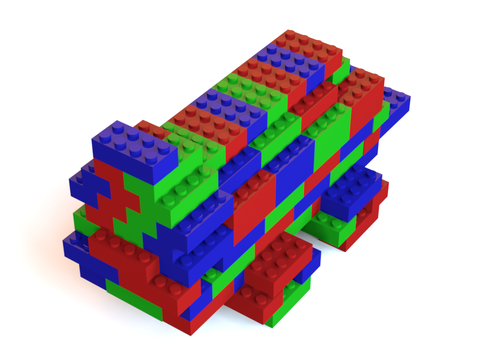}
        \includegraphics[width=0.19\textwidth]{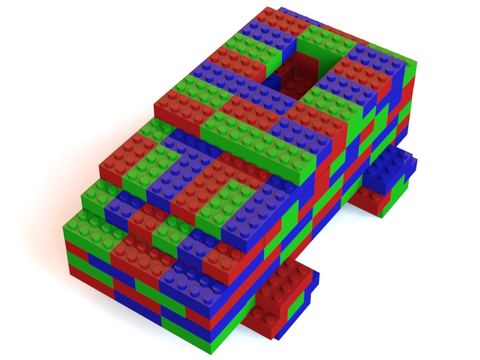}
        \includegraphics[width=0.19\textwidth]{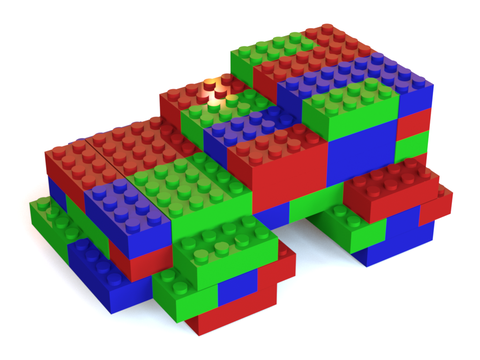}
        \includegraphics[width=0.19\textwidth]{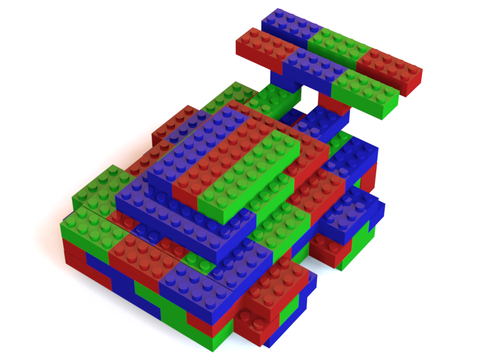}
        \includegraphics[width=0.19\textwidth]{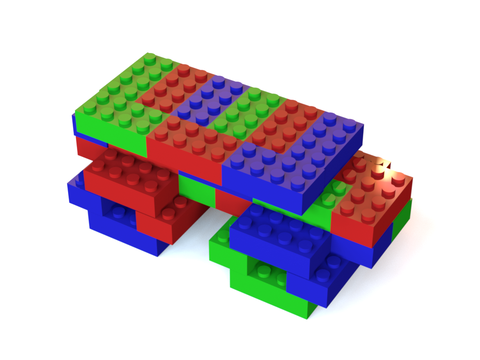}
    \caption{Examples of \emph{Car} class.}
    \label{fig:suppl_dataset_car}
\end{figure}

We demonstrate a part of our combinatorial 3D shape dataset, as shown in \figref{fig:suppl_connection_types}, and \figref{fig:suppl_dataset_bar} to \figref{fig:suppl_dataset_car}.
In addition, the statistics on our dataset is specified in \tabref{table:suppl_dataset}.

\begin{table}[t]
\centering
\caption{Statistics on combinatorial 3D shape dataset.
Std stands for standard deviation.}
\begin{tabular}{lcccccc}
    \toprule
    \textbf{Group A} & Parallel & Perpen. \\
    \midrule
    \textbf{\#instances} & 21 & 25 \\
    \textbf{Mean of \#primitives} & \multirow{2}{*}{2.0} & \multirow{2}{*}{2.0} \\
    \textbf{per instance} & & \\
    \textbf{Std of \#primitives} & \multirow{2}{*}{0.0} & \multirow{2}{*}{0.0} \\
    \textbf{per instance} & & \\
    \midrule
    \textbf{Group B} & Bar & Line & Plate & Wall & Cuboid & Sq. Pyramid \\
    \midrule
    \textbf{\#instances} & 30 & 30 & 30 & 30 & 30 & 30 \\
    \textbf{Mean of \#primitives} & \multirow{2}{*}{11.9} & \multirow{2}{*}{32.5} & \multirow{2}{*}{56.0} & \multirow{2}{*}{27.9} & \multirow{2}{*}{26.4} & \multirow{2}{*}{164.0} \\
    \textbf{per instance} & & \\
    \textbf{Std of \#primitives} & \multirow{2}{*}{6.6} & \multirow{2}{*}{42.0} & \multirow{2}{*}{35.1} & \multirow{2}{*}{14.6} & \multirow{2}{*}{17.6} & \multirow{2}{*}{129.2} \\
    \textbf{per instance} & & \\
    \midrule
    \textbf{Group C} & Bench & Sofa & Cup & Hollow & Table & Car \\
    \midrule
    \textbf{\#instances} & 30 & 30 & 30 & 30 & 30 & 30 \\
    \textbf{Mean of \#primitives} & \multirow{2}{*}{55.4} & \multirow{2}{*}{59.6} & \multirow{2}{*}{49.7} & \multirow{2}{*}{46.3} & \multirow{2}{*}{36.9} & \multirow{2}{*}{83.6} \\
    \textbf{per instance} & & \\
    \textbf{Std of \#primitives} & \multirow{2}{*}{28.0} & \multirow{2}{*}{30.5} & \multirow{2}{*}{31.2} & \multirow{2}{*}{31.8} & \multirow{2}{*}{19.2} & \multirow{2}{*}{41.0} \\
    \textbf{per instance} & & \\
    \bottomrule
\end{tabular}
\label{table:suppl_dataset}
\end{table}

\end{appendices}

\end{document}